\documentclass[lettersize,journal]{IEEEtran}
\usepackage{amsmath,amsfonts}
\usepackage{algcompatible}
\usepackage{algorithm}
\usepackage{algpseudocode}%
\usepackage{array}
\usepackage{comment}
\usepackage{textcomp}
\usepackage{stfloats}
\usepackage{url}
\usepackage{verbatim}
\usepackage{graphicx}
\usepackage{xcolor}
\usepackage{cite}
\usepackage{amsmath}
\usepackage{tabularx}
\usepackage[english]{babel}
\usepackage{amsthm}
\usepackage{multirow}
\usepackage{booktabs}
\usepackage{mathtools}
\usepackage{subcaption}
\usepackage{caption}
\usepackage{color,soul}

\usepackage[breaklinks=true]{hyperref}

\def\BibTeX{{\rm B\kern-.05em{\sc i\kern-.025em b}\kern-.08em
    T\kern-.1667em\lower.7ex\hbox{E}\kern-.125emX}}
\begin{document}

\title{
From Imperfect Signals to Trustworthy Structure: Confidence-Aware Inference from Heterogeneous and Reliability-Varying Utility Data
\\
}

\author{Haoran~Li*,~\IEEEmembership{Member,~IEEE,}
Lihao~Mai*,~\IEEEmembership{Student Member,~IEEE,}
Muhao~Guo*,~\IEEEmembership{Student Member,~IEEE,}
Jiaqi~Wu*,~\IEEEmembership{Student Member,~IEEE,}
Yang~Weng,~\IEEEmembership{Senior Member,~IEEE,}
Yannan~Sun,~\IEEEmembership{Senior Member,~IEEE,}
Ce~Jimmy~Liu,~\IEEEmembership{Member,~IEEE,}
\thanks{Haoran Li, Lihao Mai, Muhao Guo, Jiaqi Wu, and Yang Weng are with the Department of Electrical, Computer and Energy Engineering, Arizona State University, Tempe, AZ, 85281, USA. E-mail: \mbox{\{lhaoran,lmai7,mguo26,jiaqiwu1,yang.weng\}@asu.edu}. 

Yannan Sun and Ce Jimmy Liu are with Oncor Electric Delivery, Dallas, TX, 75202, USA. E-mail: \mbox{\{Yannan.Sun, Ce.Liu3\}@oncor.com}.
}

\thanks{*~The first four authors contributed equally to this work.}
\vspace{-10mm}
}

\maketitle

\begin{abstract}
Accurate distribution grid topology is essential for reliable modern grid operations. However, real-world utility data originates from multiple sources with varying characteristics and levels of quality. In this work, developed in collaboration with Oncor Electric Delivery, we propose a scalable framework that reconstructs a trustworthy grid topology by systematically integrating heterogeneous data. We observe that distribution topology is fundamentally governed by two complementary dimensions: the spatial layout of physical infrastructure (e.g., GIS and asset metadata) and the dynamic behavior of the system in the signal domain (e.g., voltage time series). When jointly leveraged, these dimensions support a complete and physically coherent reconstruction of network connectivity. To address the challenge of uneven data quality without compromising observability, we introduce a confidence-aware inference mechanism that preserves structurally informative yet imperfect inputs, while quantifying the reliability of each inferred connection for operator interpretation. This soft handling of uncertainty is tightly coupled with hard enforcement of physical feasibility: we embed operational constraints—such as transformer capacity limits and radial topology requirements—directly into the learning process. Together, these components ensure that inference is both uncertainty-aware and structurally valid, enabling rapid convergence to actionable, trustworthy topologies under real-world deployment conditions. The proposed framework is validated using over 8000 meters' data on 3 feeders in Oncor's service territory, demonstrating over 95\% accuracy in topology reconstruction and substantial improvements in confidence calibration and computational efficiency relative to baseline methods.
\end{abstract}



\vspace{-2mm}

\section{Introduction}
\label{sec:intro}

Accurate knowledge of distribution system topology is foundational to modern grid operations, yet in practice, many utilities face challenges in maintaining accurate records \cite{li2021distribution, ma2023hd}. 
Our collaboration with Oncor, one of the largest and fastest-growing electric utilities in the United States \cite{Oncor2025Q1}, is motivated by persistent challenges in maintaining accurate distribution connectivity models. These challenges stem from high-volume field activities, new design projects added to the system, and occasional human errors that require greater academic attention \cite{weng2016distributed}. For instance, emergency restoration during major storms could cause temporary connectivity model changes and leave inconsistencies in the record system, which could have cascading impact on future work. These inaccuracies degrade the performance of outage localization, voltage regulation, load forecasting, and fault analysis. On the other hand, a reliable connectivity model is the key to enable advanced metering infrastructure (AMI), distributed energy resources (DERs), and automation technologies as they become more deeply integrated into the grid \cite{hosseini2020machine, mai2025guaranteed, tajer2021advanced}. 

While many academic methods have proposed data-driven solutions to this problem, our field experience indicates that these approaches often fail under real-world conditions. This is not because the algorithms are flawed, but because their assumptions diverge from utility realities. Specifically, most methods assume access to clean measurements, consistent metadata, and idealized physical models, while ignoring operational constraints, data imperfections, and the combinatorial nature of inference at scale. Motivated by these findings, we revisit the model formulation of topology identification to align with the physical, informational, and computational constraints observed in practice.

From a modeling perspective, industrial needs solutions that examine the role of sufficient input information, robust physical constraints and uncertainty quantification. Based on such a principle, we observe that traditional academic approaches often misrepresent each of these elements when applied to real-world systems \cite{oikonomou2022core}. 
First, utilities often have access to multiple data sources—particularly spatial infrastructure records (e.g., GIS coordinates and asset metadata) and signal-based measurements (e.g., voltage time series from AMI). These two modalities are not merely complementary but also sufficient to identify grid topology. this is because one reflects physical proximity while the other captures operational coupling. However, many existing approaches rely exclusively on signal data or use spatial records only for post hoc filtering, rather than treating both as first-class inputs for inference \cite{geth2023data, wang2017efficient}.

Second, while integrating more data should improve coverage and robustness, it also introduces a critical modeling challenge: heterogeneous data quality. Traditional methods tend to discard measurements or records that fail quality thresholds (e.g., based on statistical outlier detection or Chi-square residuals). But in real-world utility environments, removing lower-confidence data can severely compromise observability. This is particularly true in sparse or poorly instrumented regions. Instead, we argue for retaining structurally useful but imperfect inputs and assigning them confidence scores that quantify estimation reliability. This approach preserves system-wide observability while allowing operators to interpret and selectively trust the inferred topology \cite{wang2016phase, hosseini2020machine}.

Third, increased data inclusion and soft uncertainty modeling must be grounded by hard feasibility guarantees. We address this by embedding operational constraints, such as transformer capacity, phase balance, and radial or mesh topology, directly into the inference process. These constraints reduce the size of the solution space, eliminate implausible configurations early, and accelerate convergence to physically meaningful results. Crucially, this constraint-driven pruning complements our confidence-aware framework: while confidence captures what is likely given the data, constraints enforce what is possible given the system \cite{frank2016introduction, li2021physical, mortlock2024adaptive}.

To bridge the gap between theoretical innovation and operational deployment, we propose a new paradigm for topology identification. 
First, we introduce constraint-guided learning, which embeds operational limits, such as transformer ratings and feeder capacities, directly into the inference process to eliminate physically implausible configurations and accelerate convergence \cite{li2021physical, mortlock2024adaptive}. Second, instead of removing noisy or partially incorrect information, such as imprecise spatial coordinates, we treat them as uncertain inputs that still carry useful structural information. By incorporating this data probabilistically, the model can utilize weak signals without relying on perfect accuracy
\cite{hosseini2020machine, wang2016phase}. Third, we leverage multi-source data fusion, aligning diverse sources including geographic information, voltage time series, and asset metadata to enable cross-validation and decomposition \cite{weng2016distributed, ma2023hd}. Such a design improves robustness and interpretability. These three principles represent a shift away from purely statistical correlation methods toward physically grounded, utility-informed, and deployment-ready solutions. 


From the industrial perspective, this paper makes three primary contributions to the field of data-driven power system modeling. First, 
we present a real-world demonstration of how academic methods can be adopted by utilities to overcome practical data challenges within large-scale and complex record systems. Second, it provided a generalizable framework that integrates physical constraints, noisy metadata, and multi-source information, offering a robust blueprint for utilities facing similar data integrity challenges \cite{weng2016distributed, mortlock2024adaptive}. Third, through extensive experiments using data from over 4 millions AMI meters, we showed substantial improvements in topology accuracy, phase correctness, and confidence quantification-highlighting the proposed approach’s scalability, modularity, and deployment readiness \cite{wang2017efficient, wang2024joint}. More broadly, this work demonstrated that flawed and inconsistent data can be transformed into trustworthy system models—not by assuming data perfection, but through principled, constraint-informed learning grounded in physical reality.

In the experiments using 8331 AMI meters' data from three feeders that represent diverse setup in Oncor's territory, our method achieved over 95\% accuracy in topology reconstruction compared to baseline methods, even in the presence of location errors and noisy measurements. The scalable design reduced computational time by over 70\% when compared to existing joint estimation frameworks, demonstrating practical feasibility for field deployment across wide service territories.

The rest of this paper is organized as follows. Section II introduces Oncor's distribution systems and provides a data overview. Section III describes industrial motivation. 
Section IV presents the proposed method. 
Section V reports experimental results on real feeders. 
Section VI discusses 
lessons learned, and Section VII concludes the paper. 

\section{System Description and Problem Formulation}

\label{sec:system}

Oncor Electric Delivery operates one of the largest electric transmission and distribution systems in the United States with more than 210,000 miles of transmission and serving over 13 million customers across Texas. The system covers a diverse range of regions, including major metropolitan areas like Dallas-Fort Worth (DFW) and resource-driven regions such as the Permian Basin. Oncor's network complexity stems from the combination of legacy infrastructure, active grid modernization initiatives, and a high annual system growth rate of over 2\%.
To support distribution grid operations, Oncor maintains a comprehensive data ecosystem. This includes an integration of the distribution Geographic Information System (GIS), Outage Management System (OMS), AMI system, and the customer billing system (CCB). The connectivity information and device phases that this paper will discuss are in the GIS data. The AMI system collects measurements such as voltage magnitudes and energy (kWh) consumption at 15-minute intervals. The AMI meters are geo-tagged with both address-based and coordinate-based location data, which are migrated from CCB, though inconsistencies in location info exist due to manual entry, and synchronization gaps between systems.

\begin{figure}[htbp]
    \centering
    \begin{subfigure}[b]{0.15\textwidth}
        \centering
        \includegraphics[width=\textwidth]{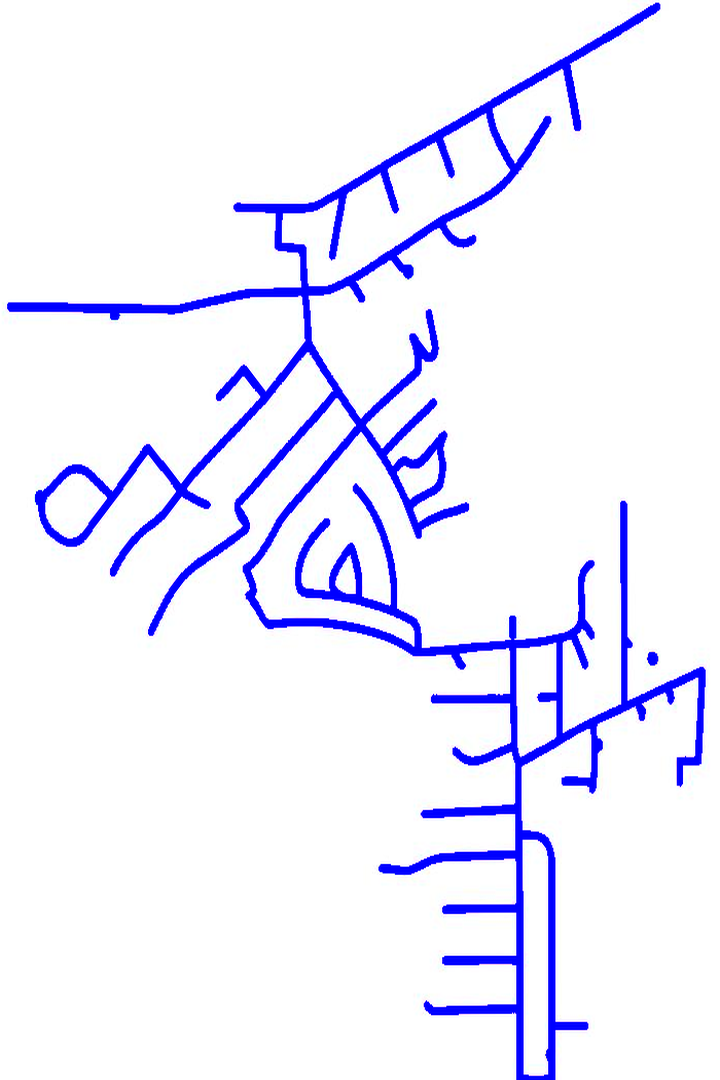}
        \caption{Feeder 1}
        \label{fig:feeder_1}
    \end{subfigure}
    \hfill
    \begin{subfigure}[b]{0.15\textwidth}
        \centering
        \includegraphics[width=\textwidth]{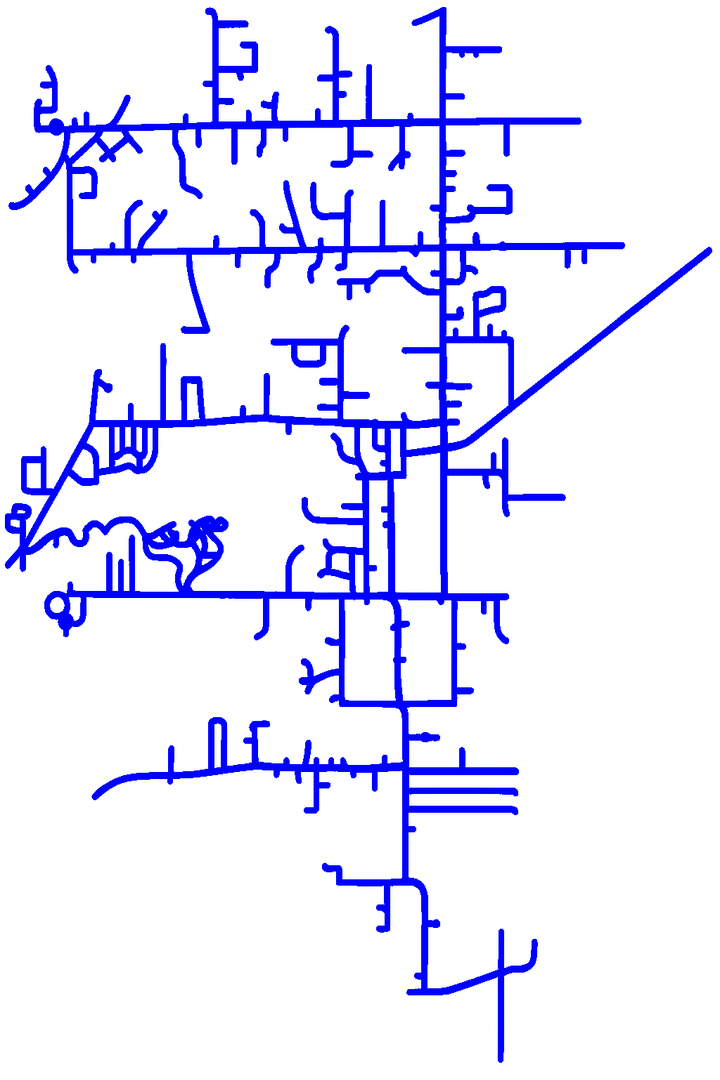}
        \caption{Feeder 2}
        \label{fig:feeder_2}
    \end{subfigure}
    \hfill
    \begin{subfigure}[b]{0.15\textwidth}
        \centering
        \includegraphics[width=\textwidth]{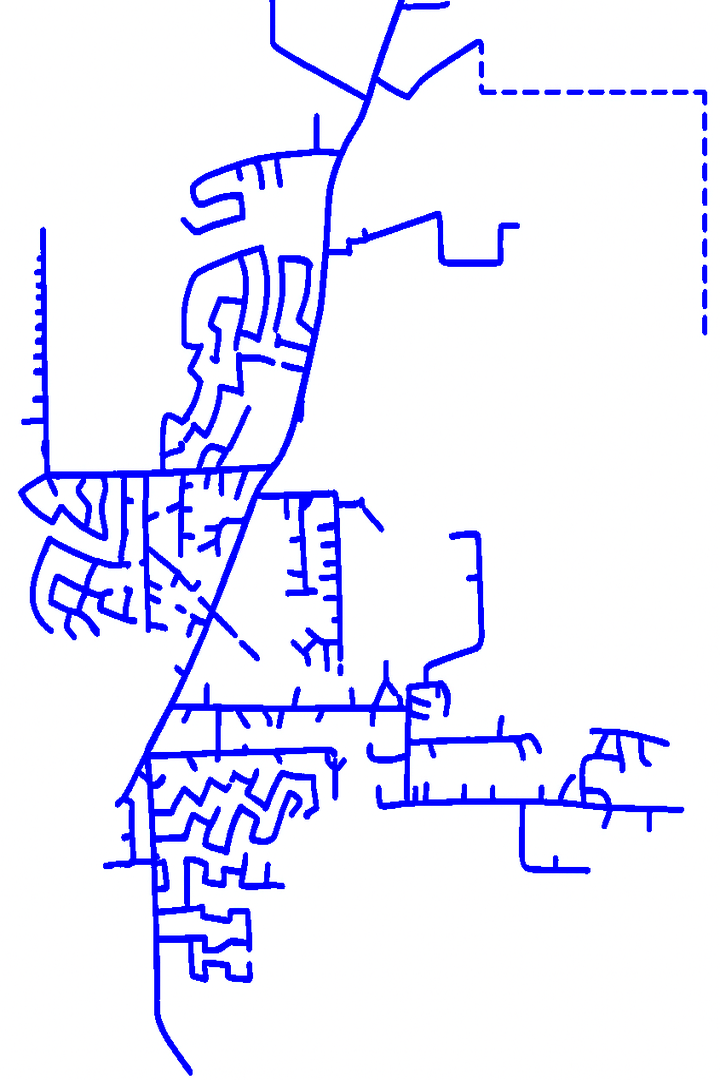}
        \caption{Feeder 3}
        \label{fig:feeder_3}
    \end{subfigure}

    \caption{Feeders in Texas provided by Oncor}
    \label{fig:feeders}
\end{figure}

Figure \ref{fig:feeders} illustrates three distribution feeder topologies provided by Oncor. 
The leftmost feeder is the smallest, serving approximately 100 transformers with 1,400 premises. The middle feeder is medium-sized, with roughly 1,500 transformers and 3,700 premises. The rightmost feeder is the largest, consisting of about 1,000 transformers but serving approximately 5,800 premises. The following is the problem definition in this utility work. 
%

\begin{itemize}
\item Problem: Large-scale topology for distribution systems.
\item Input: 
%
\textit{Geographical data}: 
%
$\phi_{i}$: Latitude of node $i$, 
$\lambda_{i}$: Longitude of node $i$.
\textit{Electrical measurements}: 
$\{V^{A}_{it}\}_{t}$: Average voltage magnitude of phase A at node $i$ and time $t$, computed over a 15-minute interval. Similar definitions apply for phases B and C.
$\{P_{it}\}_{t}$: Total active power at node $i$ and time $t$, aggregated over a 15-minute interval, measured in kilowatts (kW).
$\{Q_{it}\}_{t}$: Total reactive power at node $i$ and time $t$, aggregated over a 15-minute interval, measured in kilovolt-amperes reactive (kVAR). 
\textit{Text data:} $S_{i}$: Address string for node $i$. \textit{Physical parameters:} $V^{A}_{\mathrm{nom}}$: Nominal voltage magnitude for phase A. Similar notations for phases B and C. $\overline{S}_{j}$: Rated apparent power (capacity) of the $j$-th transformer, measured in kilovolt-amperes (kVA). $\mathcal{C}$: the set of physical constraints. For example, \textit{Voltage range}: \(V^{A}_{it} \in [  V_{\mathrm{nom,min}}^{A}, \, V_{\mathrm{nom,max}}^{A}]\). \textit{Transformer capacity constraint}: \(\sum_{i \in \mathcal{N}(j)} (P_{it}  + j Q_{it}) \le \overline{S}_{j}\), where $\mathcal{N}_j$ is the set of load nodes connected to the transformer. \textit{Base topology and phase information}: 
$\mathcal{G}_{B}=\{\mathcal{V},\mathcal{E}_B,\Phi_B\}$, where $\mathcal{V}$ is the node set, $\mathcal{E}_B$ is the base edge set, and $\Phi_B$ is the base phase set. 
$\Phi_B$ is the set of phase data for $\forall i\in\mathcal{V}$, i.e., $\Phi_B=\{\Phi^j_{i}|\Phi^j_{i}=1,j\in\{A,B,C\}\}_{i\in\mathcal{V}}$, where $\Phi^A_{i}=1$ implies that node $i$ has phase A. Similar notations for phases B and C. The base information is gathered from Utility's profiles with around $80\%$ accuracy compared to the ground truth. 

\item Output: The topology  estimation $\hat{\mathcal{G}}=\{\mathcal{V},\hat{\mathcal{E}},\hat{\Phi}\}$.

\end{itemize}

\color{black}



\section{Proposed model}

The proposed framework consists of three sequential stages. The first stage focuses on identifying meter outliers that could compromise downstream analysis. To do this, we use a combination of methods: (1) checking for inconsistencies between recorded GPS coordinates and address strings to detect spatial mismatches; (2) analyzing voltage data to flag abnormal patterns; and (3) incorporating feedback from other processes, such as intermediate phase identification results or expert review. This stage helps filter out problematic meters while preserving system observability. The second stage performs meter-to-transformer correction using machine learning models that are guided by physical constraints, such as transformer capacity limits and voltage similarity among connected meters. This ensures physically plausible and operationally realistic connections \cite{li2021physical, mortlock2024adaptive}. In the third stage, we carry out phase identification and correction by combining supervised and unsupervised learning methods. Rather than relying solely on clean data, this stage incorporates noisy or partially incorrect labels, such as mismatched phase information, as probabilistic inputs. This allows the model to cross-validate and recover accurate phase assignments \cite{wang2016phase, hosseini2020machine}.

\label{sec:model}

\subsection{Data Preprocessing}

\begin{figure}[h!]
\centering
\includegraphics[width=\columnwidth]{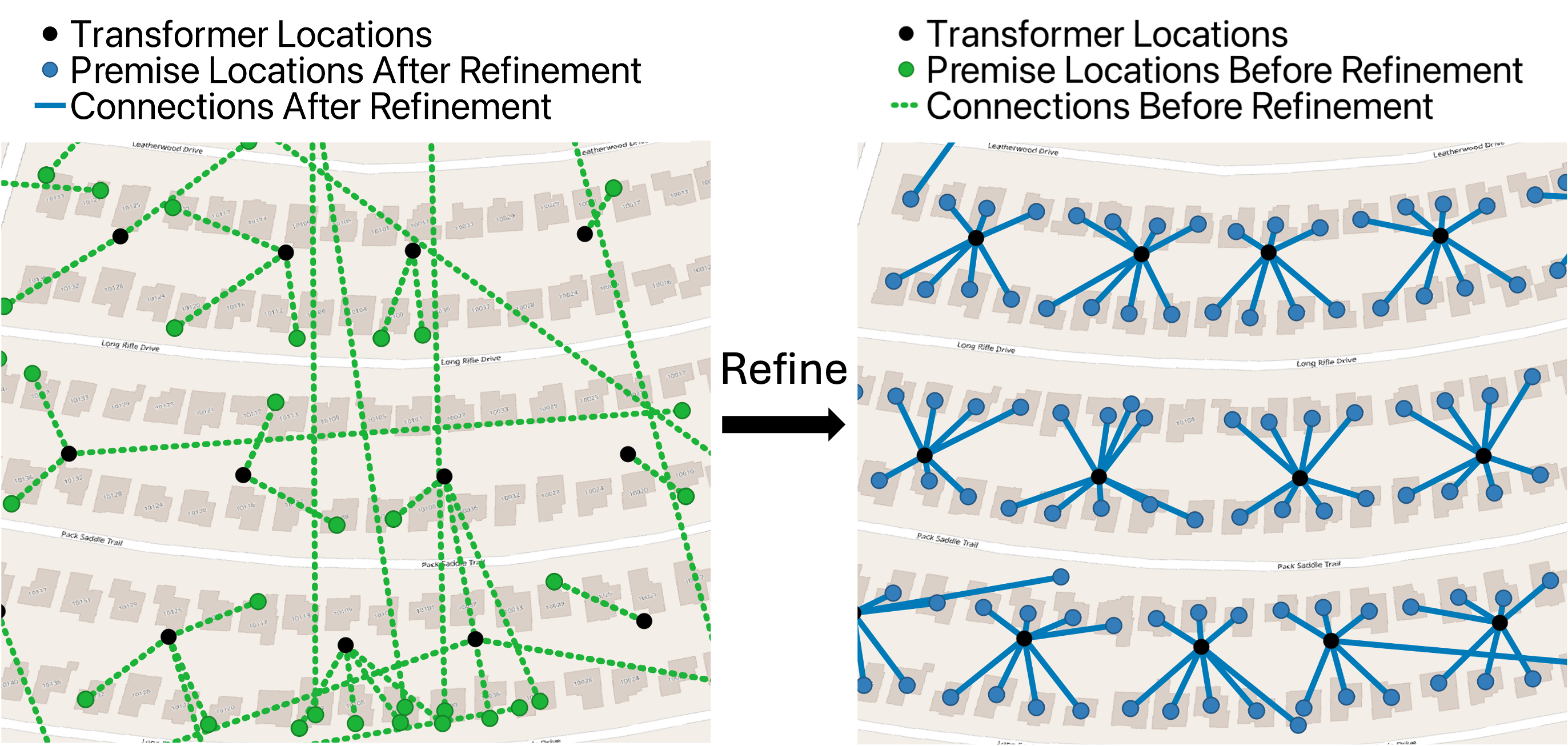}
\centering
\caption{Before and after geocoding-based refinement.}
\label{fig:geocoding}
\end{figure}

The input datasets defined in Section~\ref{sec:system} often suffer from spatial inconsistencies. For example, customer-to-transformer mappings are typically derived from legacy records or billing systems, which may contain outdated or erroneous address information. These discrepancies are especially prevalent in utilities like Oncor, where field inspection reports have identified customer meters linked to transformers that are geographically implausible, sometimes located miles apart or across natural barriers like highways or rivers.
%
%
 Figure~\ref{fig:geocoding} illustrates a representative case using Oncor’s data, where several premises appear to be connected to a transformer that is not the geographically closest. These inaccurate associations result from erroneous or outdated latitude and longitude records linked to either the premises or the transformers. To mitigate this issue, we apply geocoding, which converts address-level information, such as street address, city, and ZIP code, into precise geographic coordinates. This process enhances the spatial accuracy of the data and ensures that each asset is correctly located. 

Specifically, we apply geocoding, the process of converting address-level information, e.g., street names, cities, and ZIP codes, into geographic coordinates, i.e., latitude and longitude, enabling spatial reasoning about the physical layout of customer premises and transformer assets. We use open-source platforms such as Nominatim, based on OpenStreetMap, and commercial APIs like Google Maps Geocoding API. They take structured address data, such as city, ZIP code, street address, and optionally transformer latitude and longitude as input and return standardized geolocation outputs. Such output includes coordinates, address quality, and administrative boundaries \cite{clemens2015geocoding}. These geocoded results allow us to verify the consistency between customer locations and their assigned transformers. We can also detect mismatches where customers are located outside the expected service area and quantify spatial discrepancies based on the distance between each premise and its associated transformer. 
\begin{figure}[h!]
\centering
\includegraphics[width=\columnwidth]{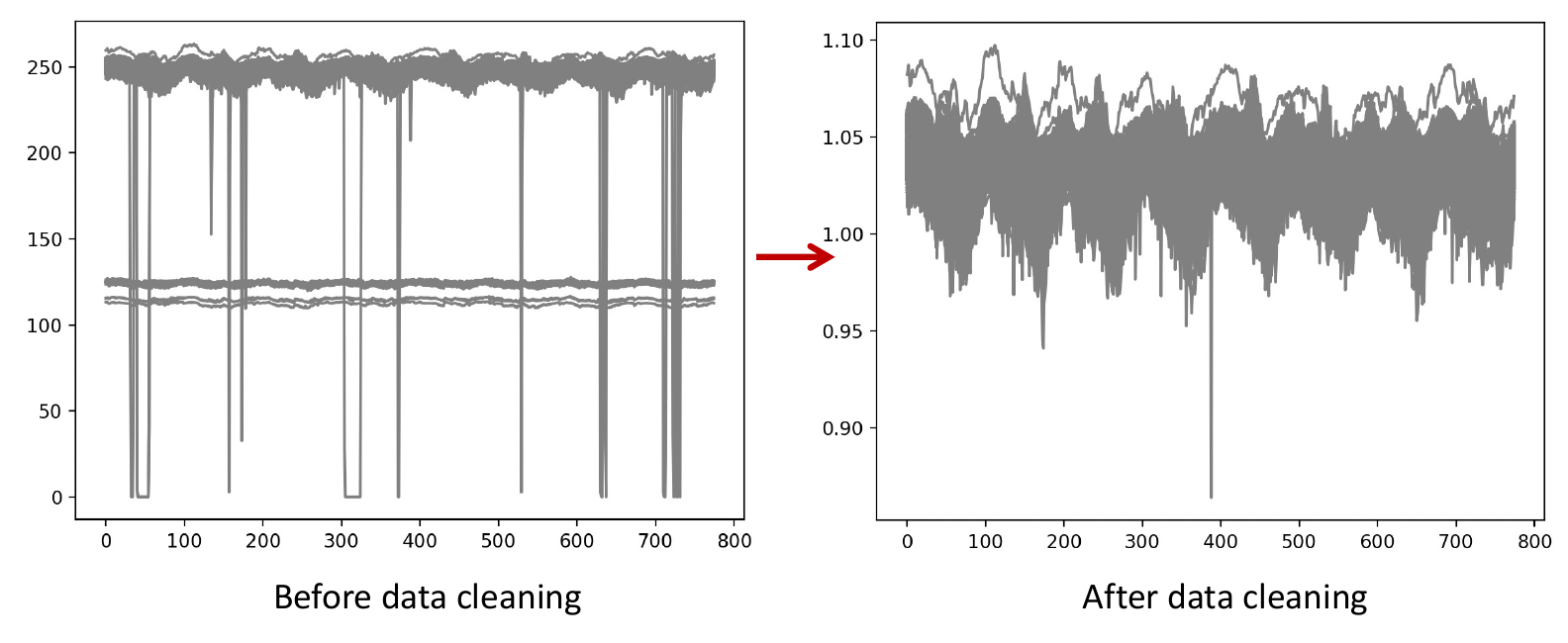}
\centering
\caption{The voltages before and after data cleaning.}
\label{fig:data_filtering}
\end{figure}

A second data quality issue comes from the (1) electrical measurements, including missing nominal voltage labels, especially for some premises, (2) incomplete time-series readings, which impair temporal correlation analysis, and (3) and unrealistic measurements, such as persistently low or saturated voltage values that deviate from physical plausibility. For example, in Oncor’s feeder data, some meters report sustained voltage values far below expected operating levels (e.g., under $100V$). This indicates potential sensor faults or logging errors.
For data quality issues, we apply data imputation processes. Such processes include voltage normalization, where missing nominal voltage levels are inferred from neighboring premises and used to standardize voltage readings. Records with incomplete time-series data are filtered out to preserve consistency in temporal analyses. Outliers with extreme voltage values are removed using the interquartile range (IQR) \cite{vinutha2018detection} method, and unrealistic patterns such as flatlined voltage profiles are discarded. In Figure \ref{fig:data_filtering}, we show the visualization of voltage before and after the data cleaning.


\vspace{-3mm}
\subsection{Outlier Detection Using Fused Datasets}
\label{shapeaware}

\begin{figure}[h!]
\centering
\includegraphics[width=1\linewidth]{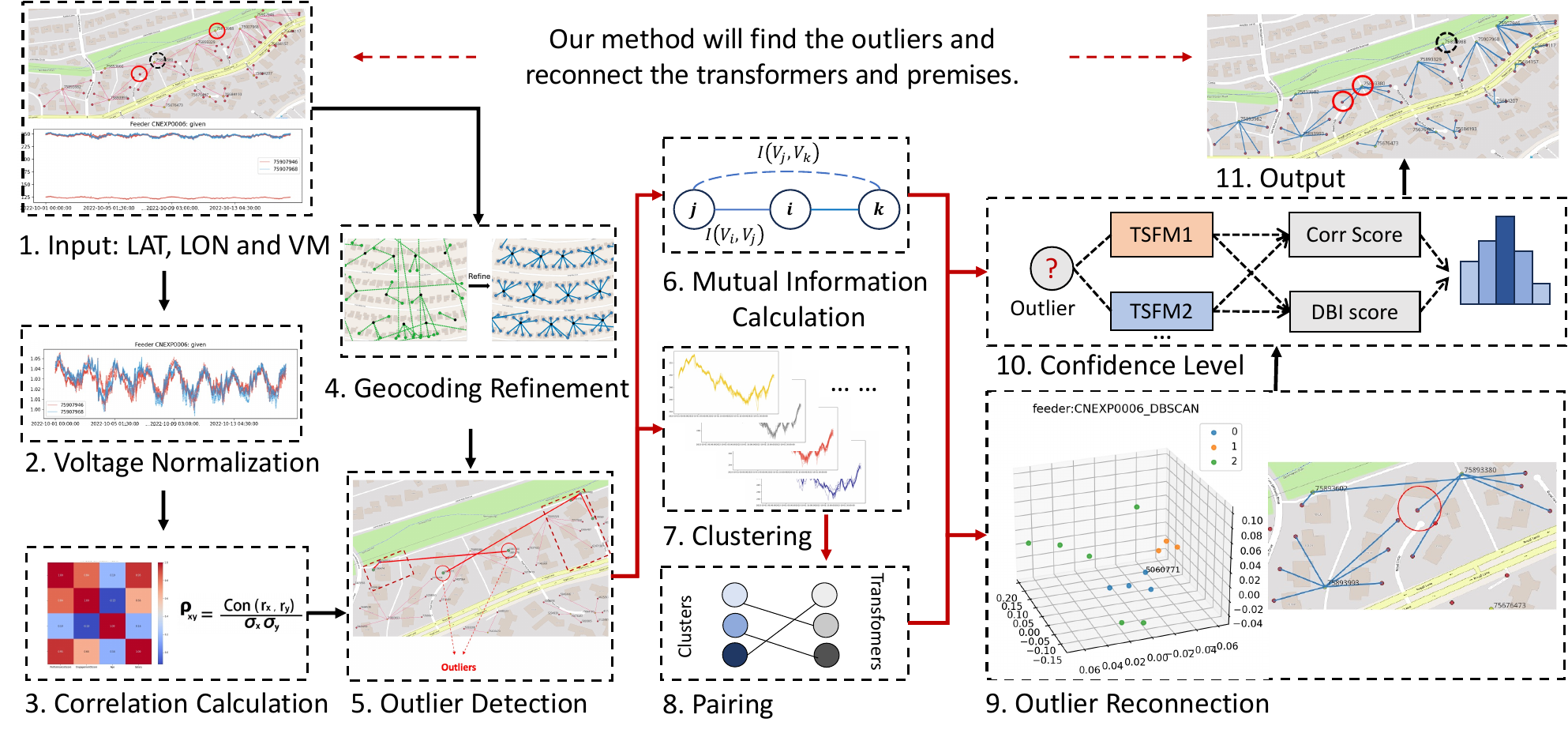}
\centering
\caption{Flowchart of the topology identification process. 
}
\label{fig:flowchart}
\end{figure}


The workflow in Figure~\ref{fig:flowchart} begins by inputting refined geographical and electrical datasets, including location coordinates (LAT, LON) and voltage magnitude (VM) measurements (Step~1). The voltage data are then normalized (Step~2), and correlation calculations are performed (Step~3) to reveal electrical relationships among premises. A geocoding refinement (Step~4) integrates the spatial layout with voltage correlations to detect abnormal connections and isolate suspicious nodes as outliers (Step~5). To enhance this detection, mutual information between nodes is computed (Step~6), and clustering is applied to group consistent connections (Step~7). The algorithm then pairs valid premises and transformers (Step~8) and reconnects isolated outliers within their local neighborhoods (Step~9). Finally, each connection’s confidence level is evaluated (Step~10) using correlation and clustering scores, producing a verified topology output (Step~11).

Underlying this workflow is a two-level divide-and-conquer strategy that makes large-scale topology identification computationally tractable. In the first level, geographical proximity and electrical correlations are used to identify abnormal connectivity and separate outlier nodes based on the base topology~$\mathcal{G}_{B}$. In the second level, the reconnection process updates the base topology by resolving the outlier nodes within their surrounding regions.
To support this robust detection, our method explicitly leverages multi-source data fusion. Instead of relying on a single type of measurement, the framework aligns diverse sources, geocoded address information, voltage time-series patterns, and asset metadata, to cross-validate connectivity from different perspectives. By integrating these heterogeneous signals, the model achieves stronger outlier detection and greater scalability than using electrical or geographical data alone.

Mathematically, we employ the geocoded coordinates to verify the physical feasibility of premise-to-transformer connections. For each premise~$i$ (household node~$i$) with coordinates $(\phi_i, \lambda_i)$ and its assigned transformer~$j$ with coordinates $(\phi_j, \lambda_j)$, the geodesic distance is computed as
\begin{equation}
\label{eqn:geo_dis}
d_{i,j} = \text{GeoDist}\big((\phi_i, \lambda_i), (\phi_j, \lambda_j)\big),
\end{equation}
where $\text{GeoDist}(\cdot, \cdot)$ denotes the geodesic distance between two points on the Earth’s surface, computed from their latitude and longitude coordinates using a spherical or ellipsoidal model of the Earth \cite{karney2013algorithms}. We also compute the distance to the closest transformer in the service area:
$d_{i,j^*} = \min_{k} \text{GeoDist}\big((\phi_i, \lambda_i), (\phi_k, \lambda_k)\big)$,
where $k$ indexes all transformers. The \textit{distance ratio} is then defined as
$r_i = \frac{d_{i,j}}{d_{i,j^*}}$.
A large value of $r_i$ suggests that premise~$i$ may be incorrectly assigned. We flag it as a \textit{geographic outlier} if $r_i > \tau$, where $\tau$ is a user-defined threshold.
To further validate suspicious premise-to-transformer assignments, we incorporate electrical correlation analysis using voltage measurements among premises connected to the same transformer. Let $j$ denote a transformer index, and let $\mathcal{N}(j)$ be the set of all premises connected to transformer $j$ under the base topology $\mathcal{G}_B$. For each premise $i \in \mathcal{N}(j)$, let $\{V_{it}\}_t$ denote the time series of voltage magnitudes at premise $i$. For any pair of premises $i_1, i_2 \in \mathcal{N}(j)$, we compute the Pearson correlation coefficient
$\rho_{i_1, i_2} = \text{Corr}\left(\{V_{i_1 t}\}_t, \{V_{i_2 t}\}_t\right)$,
based on their voltage time series.
For each premise $i \in \mathcal{N}(j)$, we evaluate its electrical consistency with the rest of the group. If
$\rho_{i,k} < \epsilon \quad \text{for all } k \in \mathcal{N}(j),\ k \ne i$,
then premise $i$ is considered \textit{electrically uncorrelated} with all other premises assigned to transformer $j$. In this case, the connection between premise $i$ and transformer $j$ is flagged as \textit{suspicious}, indicating a likely topology error.

\vspace{-3mm}
\subsection{ML-based Reconnection and Topology Refinements}

\label{sec:recon}
\begin{figure}[h!]
\centering

\includegraphics[width=1.08\linewidth]{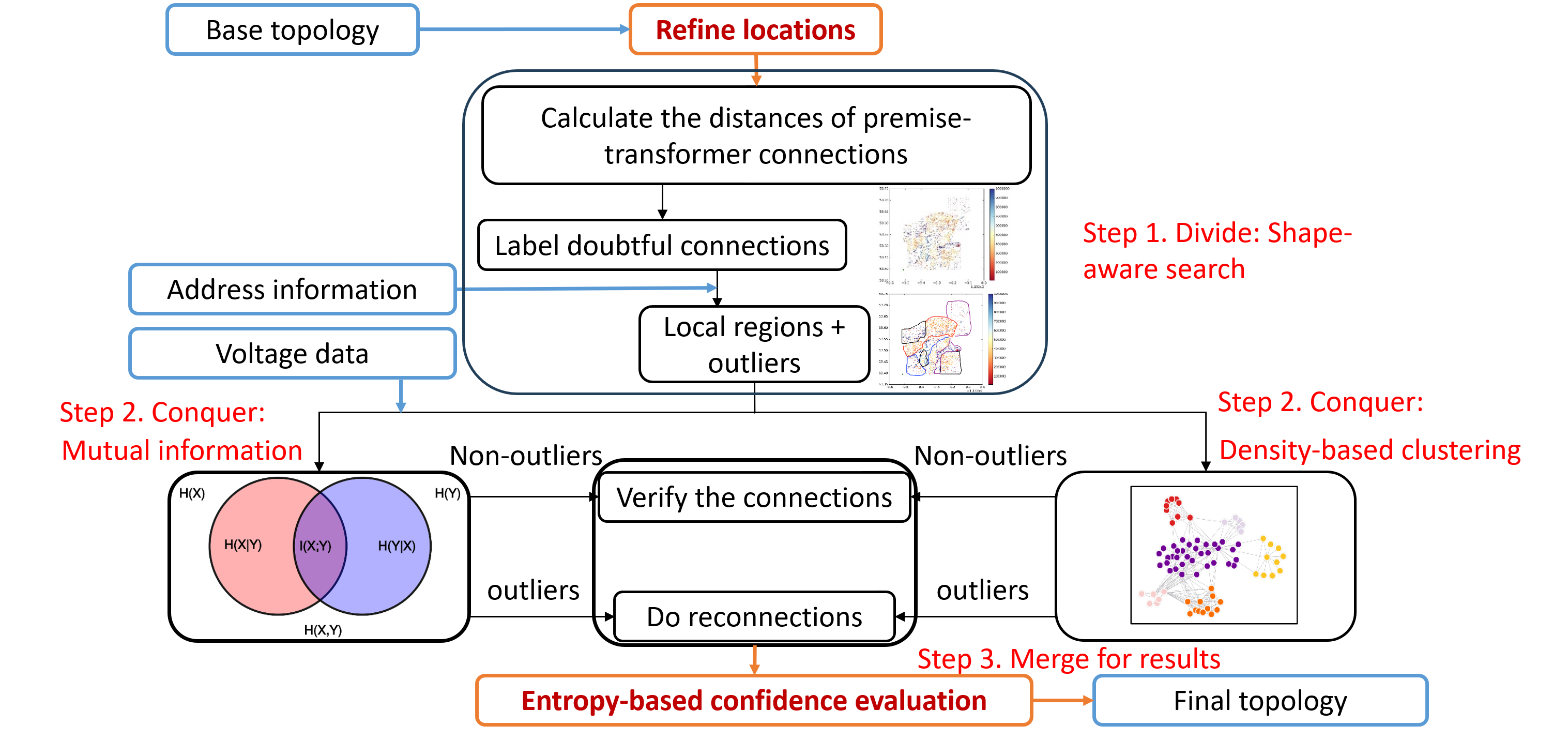}
\centering
\caption{Framework of the topology identification. 
\label{fig:Frameworkorigin}
}
\end{figure}
\label{sec:model}




After detecting anomalous transformer-to-premise connections, it is essential to reconnect the outlier nodes to their correct transformers. This reverse-engineering task aims to recover hidden physical connections based on operational data. While previous data-driven methods~\cite{weng2016distributed,weng2016distributed2,li2021distribution,li2023distribution} have addressed similar problems, they often rely solely on electrical measurements and must infer the entire topology from scratch, which limits their scalability for very large-scale feeders.

In contrast, Figure~\ref{fig:Frameworkorigin} shows our framework that leverages both the geographical data and an initial base topology~$\mathcal{G}_B$. Specifically, the framework follows a divide-and-conquer strategy. In the divide step, a shape-aware search refines premise locations using address information and voltage data to detect and label suspicious connections. In the conquer step, mutual information analysis and density-based clustering verify doubtful connections and reconnect isolated outliers within their local regions. An entropy-based confidence evaluation then merges the verified and reconnected connections to output the final refined topology.

A key innovation in this process is that instead of discarding noisy or partially inaccurate information, such as imprecise coordinates or small voltage mismatches, we treat these signals as probabilistic clues that still carry useful structural information. By integrating uncertainty rather than filtering it out, the model leverages weak but relevant patterns to guide reconnection where direct measurements are insufficient or ambiguous.
%
%
Altogether, this multi-stage process ensures that after the initial anomaly detection, only the connections for outlier nodes need to be recovered. For each outlier node~$i$, the reconnection search is limited to its neighboring transformer set~$\mathcal{N}_T(i)$, which makes the problem tractable even for feeders with thousands of premises.

After detecting an outlier $i$, we gather its $k$ nearest transformers and form the candidate‑premise set  
$\mathcal{D}_i=\{j\mid\forall j\in\mathcal{N}(k),\,k\in\mathcal{N}_T(i)\}$.
For every premise $j\in\mathcal{D}_i$ we construct a joint spatial–electrical feature vector
$\mathbf{x}_j=\bigl[V^{A}_j(0),\,V^{A}_j(1),\dots,V^{A}_j(t),\;\phi_j,\;\lambda_j\bigr]\in\mathbb{R}^{t+3}$,
where $V^{A}_j(\tau)$ is the phase‑$A$ voltage at time step $\tau$, $\phi_j$ is latitude, and $\lambda_j$ is longitude.  
K‑means clustering is then used to partition these vectors into $K$ groups.  K‑means iteratively (i) assigns each point to the nearest centroid in Euclidean space and (ii) updates each centroid to the mean of the points assigned to it, repeating until the centroids no longer move. 
After clustering the candidate premises into $\{\mathcal{S}_1,\dots,\mathcal{S}_K\}$, we apply a greedy assignment to pair each cluster with the transformer that most frequently appears in that cluster.  Formally, for cluster $\mathcal{S}_\ell$ we select
$T^\star(\mathcal{S}_\ell)\;=\;\arg\max_{k\in\mathcal{N}_T(i)}
\bigl|\mathcal{S}_\ell\cap\mathcal{N}(k)\bigr|$,
i.e.\ the transformer serving the largest number of premises in $\mathcal{S}_\ell$.  The outlier $i$ is then reconnected to the transformer $T^\star(\mathcal{S}_{c(i)})$ of the cluster $c(i)$ that contains it.



An alternative reconnection method uses mutual information (MI) to evaluate the statistical dependency between voltage time series at different nodes. For each premise node $i$ and for each candidate transformer neighborhood $k \in \mathcal{N}_T(i)$, we compute the average mutual information between node $i$ and all nodes in $\mathcal{N}(k)$, the set of premises currently assigned to transformer $k$. This average is defined as:
\begin{equation}
\bar{\text{MI}}(i,k) = \frac{1}{|\mathcal{N}(k)|} \sum_{k' \in \mathcal{N}(k)} \text{MI}(V_i(t), V_{k'}(t)),
\end{equation}
where $\text{MI}(V_i(t), V_{k'}(t))$ denotes the mutual information between the voltage magnitude time series of node $i$ and node $k'$. The transformer neighborhood $k$ that yields the highest $\bar{\text{MI}}(i,k)$ is selected as the most likely correct reconnection for premise $i$.

\vspace{-3mm}
\subsection{Confidence Level}
\label{confidencelevel}

\begin{figure}[h]
\centering

\includegraphics[width=1.08\linewidth]{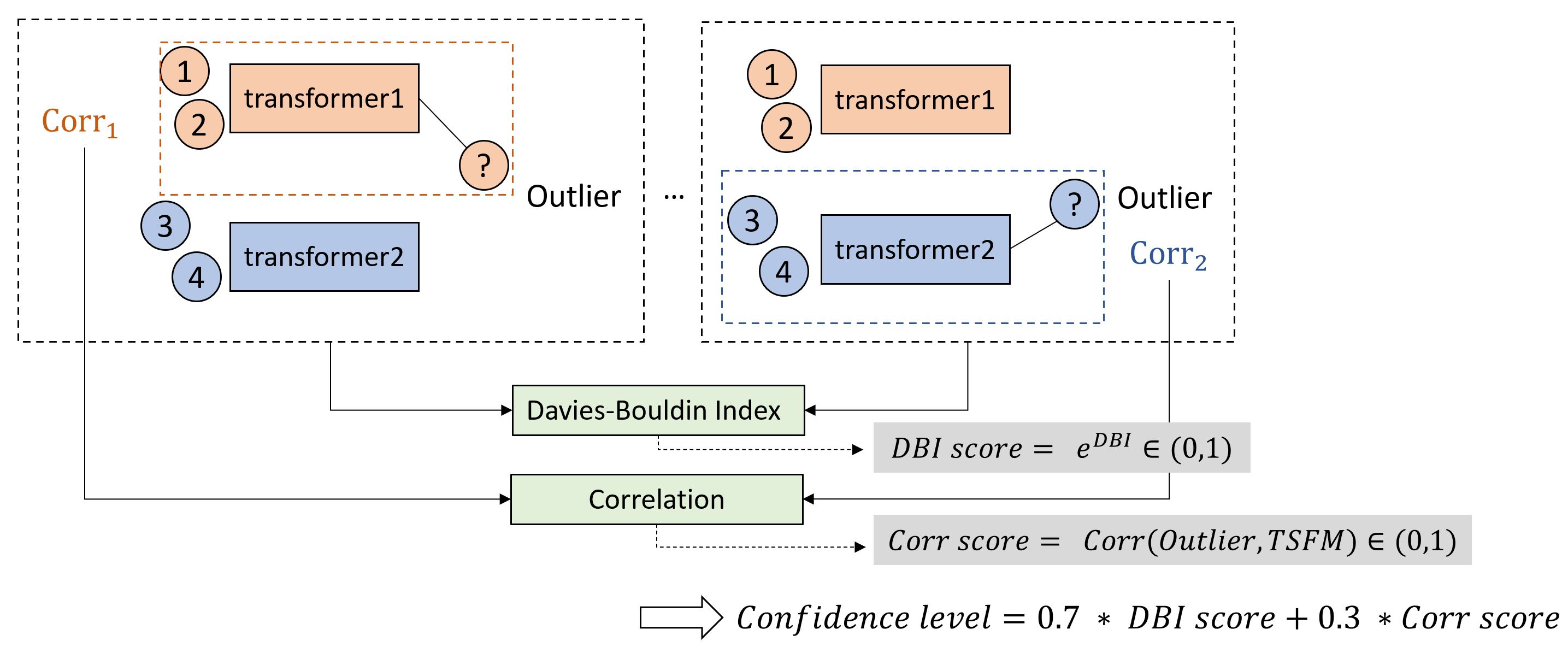}
\centering
\caption{An illustration of the confidence level design.}
\label{fig:confidence}

\end{figure}

Following topology identification, we assign confidence scores to quantify the reliability of each result, show in Figure \ref{fig:confidence}. These scores guide operational decisions and prioritize high-confidence cases when resources are constrained and flagging low-confidence cases for further investigation and algorithm refinement.

\subsubsection{Falsification as a Foundational Principle}

Falsification, as defined by \cite{popper2005logic}, refers to testing a hypothesis by actively seeking evidence that contradicts it. In our context, the hypothesis is: “The identified outlier is correctly assigned to transformer $T$.” To test this, we simulate alternate scenarios by assigning the outlier to other transformers and compare the quality of clustering and correlation.

\subsubsection{Clustering Quality via Davies-Bouldin Index (DBI)}

The Davies-Bouldin Index (DBI)~\cite{davies1979cluster} quantifies clustering quality based on intra-cluster compactness and inter-cluster separation:
%
$\text{DBI} = \frac{1}{k} \sum_{i=1}^{k} \max_{j \ne i} \left( \frac{S_i + S_j}{d(c_i, c_j)} \right)$
where $S_i$ is average distance between points in cluster $i$ and its centroid $c_i$. $d(c_i, c_j)$ is the distance between cluster centroids $c_i$ and $c_j$. For falsification application, we calculate two values. One is $\text{DBI}_{\text{true}}$, where DBI when using the outlier assignment from our model, and $\text{DBI}_{\text{false}}$, where DBI when the outlier is reassigned to another transformer (falsification). We define a normalized clustering confidence score:
$\text{Score}_{\text{DBI}} = \sigma \left( \log \left( \frac{\text{DBI}_{\text{false}}}{\text{DBI}_{\text{true}}} \right) \right), \quad \text{where } \sigma(x) = \frac{1}{1 + e^{-x}}$.
%
A higher $\text{Score}_{\text{DBI}}$ implies that the original assignment yields better clustering than any falsified alternatives.

\subsubsection{Correlation-Based Confidence}

To enhance robustness, we also analyze the correlation between the outlier and the assigned transformer. Specifically, $\text{Corr}_{\text{ours}}$ is the correlation between the outlier and the transformer assigned by our model. $\text{Corr}_{\text{falsification}}$ is the average correlation between the outlier and other transformers.
%
$\text{Score}_{\text{corr}} = \sigma \left( \frac{\text{Corr}_{\text{ours}}}{\text{Corr}_{\text{falsification}}} \right)$.
%
Because time series from transformers often share similar trends, this ratio usually lies in $(0.5, 1)$ and boosts confidence when the model’s assigned transformer shows stronger correlation.

\subsubsection{Final Confidence Level Formulation}

To fuse both clustering and similarity perspectives, we define the overall confidence level as: 
    $\text{Confidence Level} = 0.7 \cdot \text{Score}_{\text{DBI}} + 0.3 \cdot \text{Score}_{\text{corr}}$.
%
This balanced formulation ensures that both structure (via DBI) and behavior (via correlation) support the outlier's assignment.
A high confidence level (close to 1) suggests our assignment is significantly better than any falsified alternatives, both in terms of clustering coherence and statistical correlation. A lower value signals ambiguity, suggesting further validation may be needed.

\vspace{-3mm}
\subsection{Physics-Based Validation}
\label{groundtruth}

Many unsupervised or statistical identification methods rely on fixed thresholds to flag suspicious connections. However, these thresholds often miss physically implausible results when the system data are ambiguous. To overcome this, we introduce constraint-guided learning, which directly embeds operational limits, such as transformer loading bounds and feeder capacity constraints, into the validation stage.
Once the refined topology is produced, the known physical constraint set~$\mathcal{C}$ is applied to check feasibility. This physics-based validation (1) eliminates configurations that violate transformer ratings or feeder capacities, and (2) uses the same constraints to generate trust scores for all connections. If low-confidence regions are found, the topology can be iteratively adjusted and re-verified. By combining entropy-based measures with physical feasibility checks, the framework prevents unrealistic results, accelerates convergence, and ensures the final output aligns with practical grid operation limits.

For example, for transformer~$j$ with the identified connected premise set $\mathcal{N}(j)$, we compute the aggregate apparent power:
$S^{\text{agg}}_{jt} = \sum_{i \in \mathcal{N}(j)} \sqrt{P_{it}^2 + Q_{it}^2}$.
%
%
The result can be compared to the transformer’s rated capacity $\overline{S}_j$. If the aggregated power consistently exceeds $\overline{S}_j$, or if $S^{\text{agg}}_{jt}$ exhibits unrealistic fluctuations (e.g., large discontinuities or instability), then the current node-to-transformer mapping may be physically invalid.
Finally, we verify voltage compatibility between each premise and its assigned transformer by checking nominal voltage range constraints. For all $t$, the measured voltage $V^A_{it}$ at premise $i$ should lie within an acceptable operating range:
$V^A_{it} \in \left[V^{A}_{\mathrm{nom,min}}, V^{A}_{\mathrm{nom,max}}\right]$,
and similarly for other phases. Other verification constraints in $\mathcal{C}$, if provided by the utility, can also be used. 
\vspace{-3mm}
\section{Experiment}
\label{sec:experiment}




\subsection{Data Description}

The data used in this study supports comprehensive modeling of the low-voltage distribution network, capturing spatial, electrical, and temporal characteristics. This analysis focuses on three feeders ranging from compact rural configurations to dense urban layouts.
The data consists of 3 parts. The first part provides the foundational metadata for each meter. Each record includes an internal identifier (\texttt{ENDPOINTID}) and its associated transformer (\texttt{XFMR}). It also contains geolocation information for both meters and transformers. Additional attributes include phase information and nominal voltage readings across channels. Data inconsistencies might be found in this dataset such as outdated meter-to-transformer mappings, incorrect phase labels, and imprecise coordinates. These imperfections motivate preprocessing steps such as geocoding, spatial outlier detection, and voltage normalization.

The second part is the outage records on these feeders, which contain records of service interruptions. For each outage, the premise ID, outage start time, and restoration time are recorded. These events provide a temporal lens through which structural inference can be validated—by analyzing correlated outages across nearby premises, we can verify or adjust connectivity info made in the asset mapping.

The third part contains 15-minute interval voltage and load time-series data collected from the AMI meters. Each entry is associated to an internal identifier \texttt{ENDPOINTID}. 

Table~\ref{tab:feeder_stats} summarizes the scope of the three feeders studied in this project. Feeder 1 is relatively small and geographically isolated, whereas Feeders 2 and 3 cover more expansive, complex, and densely populated regions. These differences provide a robust testbed for evaluating algorithm scalability and generalizability. Figure~\ref{fig:feeders} visualizes the spatial extent and layout of these feeders.

\begin{table}[H]
\centering
\caption{Feeder Summary Statistics}
\label{tab:feeder_stats}
\begin{tabular}{c|c|c}
\toprule
\textbf{Feeder ID} & \textbf{Number of Transformers} & \textbf{Number of Premises} \\
\midrule
Feeder 1 & 115  & 1,398 \\
Feeder 2 & 1,548 & 3,677 \\
Feeder 3 & 1,046 & 5,841 \\
\bottomrule
\end{tabular}
\end{table}

While comprehensive, the datasets reflect real-world data quality challenges inherent to utility-scale deployments. Many AMI meter coordinates are outdated or imprecise, phase and transformer assignments are occasionally incorrect, and some meters lack complete voltage traces or metadata. These limitations necessitate a robust data cleaning pipeline incorporating geospatial corrections, constraint-guided validation, and statistical consistency checks to generate reliable topology and phase maps. 

\vspace{-3mm}
\subsection{Fusing Geographical and Electrical Data Mitigates Indistinguishability from Highly-Correlated Voltage}



\begin{figure}
    \centering
    \begin{subfigure}[b]{\columnwidth}
        \centering
        \includegraphics[width=0.5\columnwidth]{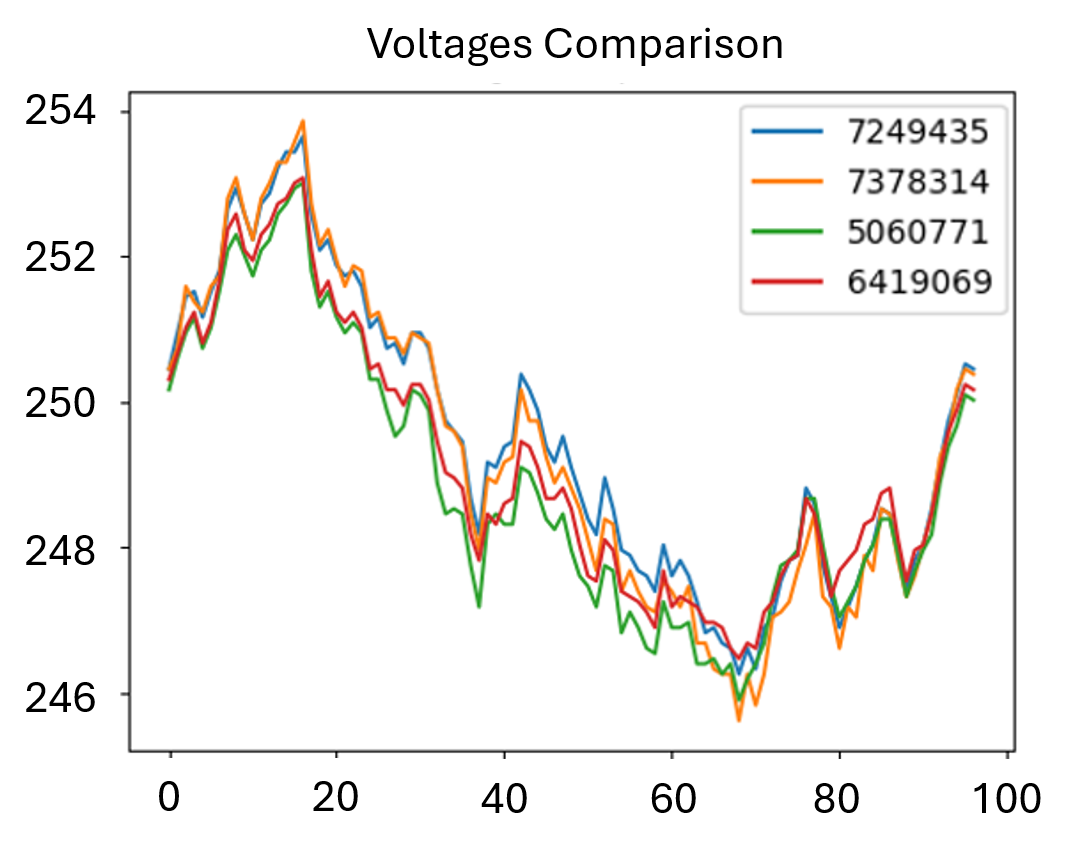}
        \caption{Between complete transformer 377108415 and 75893988.}
        \label{fig:correlation377109415and75893988}
    \end{subfigure}
    \hfill
    \begin{subfigure}[b]{0.49\textwidth}
        \centering
        \includegraphics[width=\textwidth]{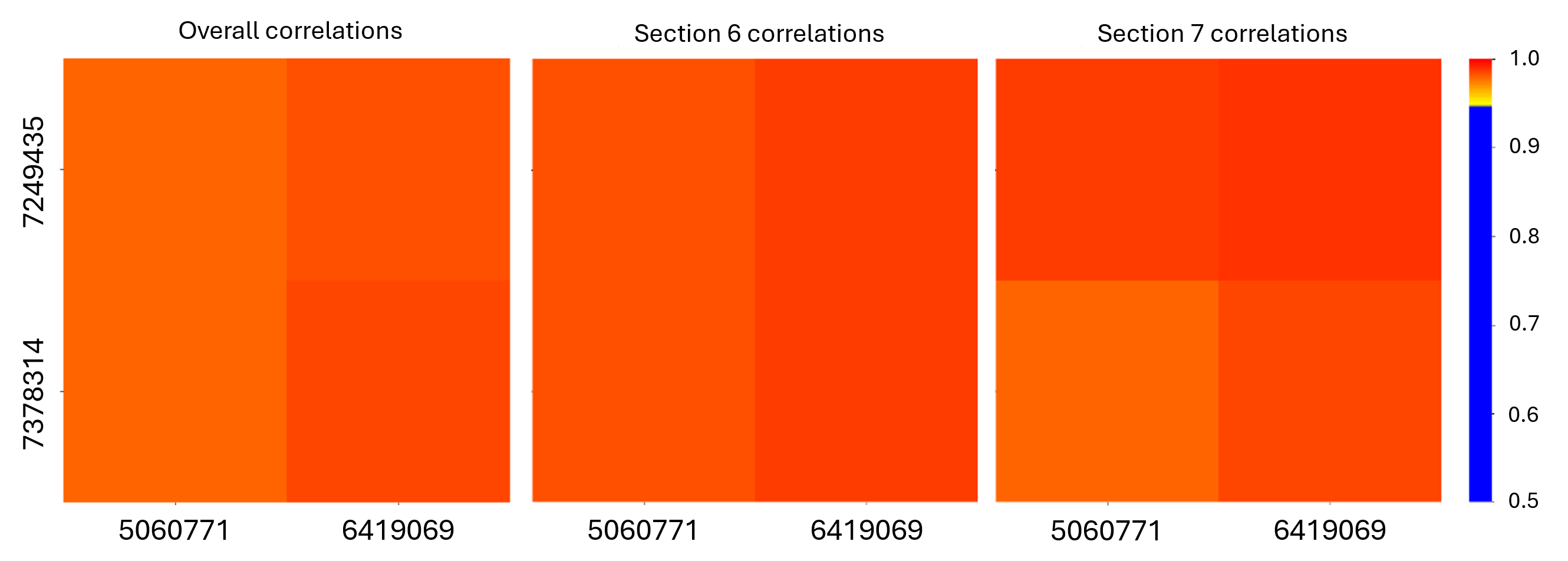}
        \caption{Correlation between transformer 377108415 and 75893988, including complete correlation and sectional correlation results.}
        \label{fig:correlationheatmapall}
    \end{subfigure}
    \caption{Correlation result of transformer 377108415 and 75893988. Transformer 377108415 contains premises 5060771 and 6419069. Transformer 75893988 contains premises 7378314 and 7249435.}
    \label{fig:Correlation test case}
\end{figure}

\begin{figure}[h]
    \centering
    \includegraphics[width=0.5\textwidth]{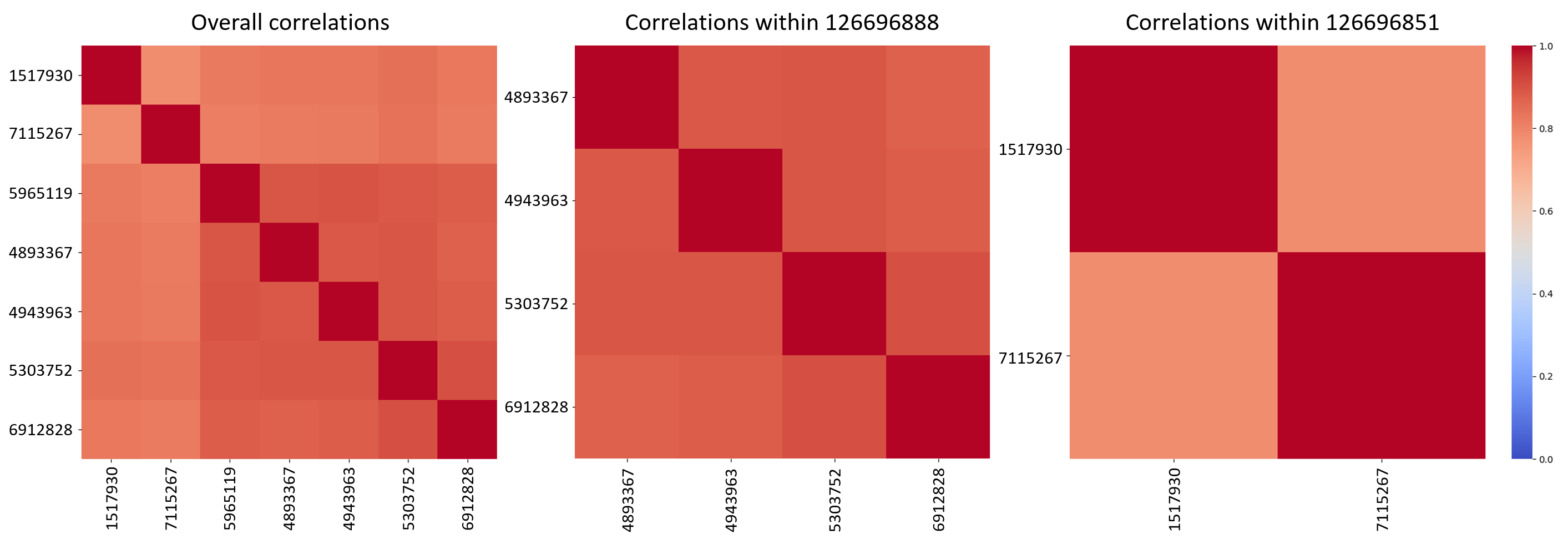}
    \caption{Among 126696864, 126696888, and 126696851. 
    }
    \label{fig:correlationheatmapallresult}
\end{figure}

\begin{figure}[htbp]
    \centering
    \begin{subfigure}[b]{0.15\textwidth}
        \centering
        \includegraphics[width=\textwidth]{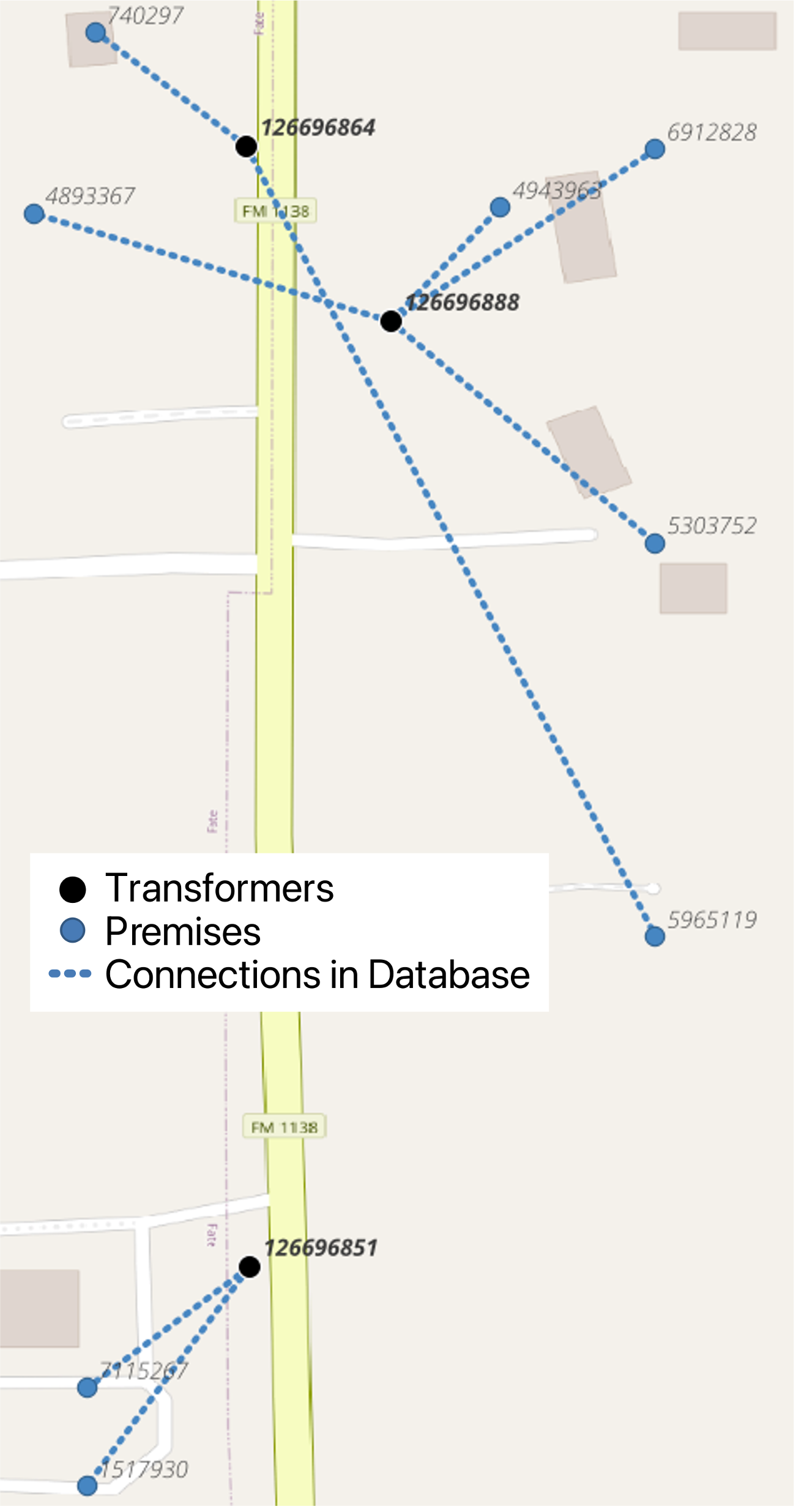}
        \caption{Connections in database.}
        \label{fig:case2_oncor}
    \end{subfigure}
    \hfill
    \begin{subfigure}[b]{0.15\textwidth}
        \centering
        \includegraphics[width=\textwidth]{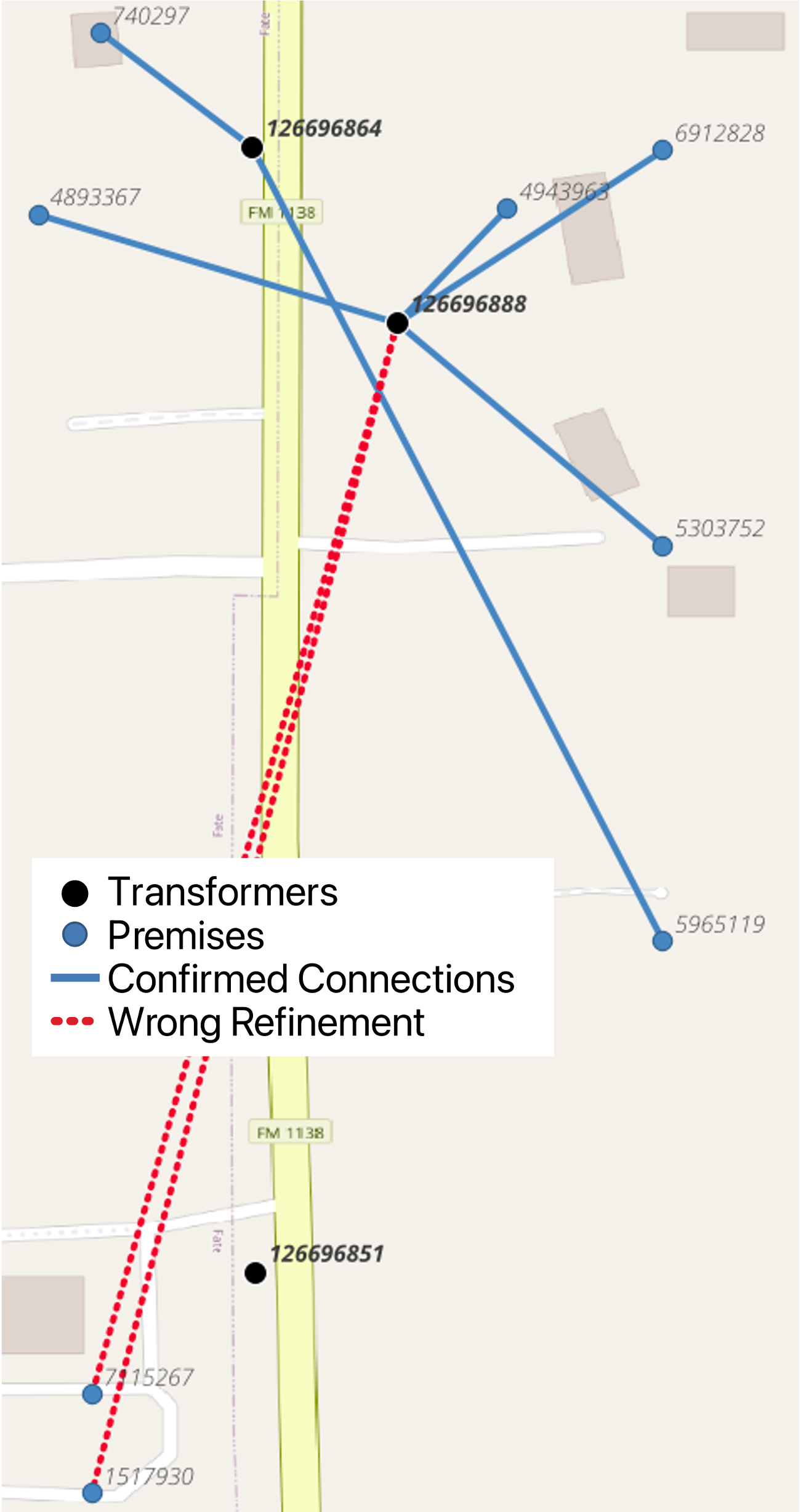}
        \caption{Bad refinement by MI only.}
        \label{fig:case2_fail}
    \end{subfigure}
    \hfill
    \begin{subfigure}[b]{0.15\textwidth}
    \centering
        \includegraphics[width=\textwidth]{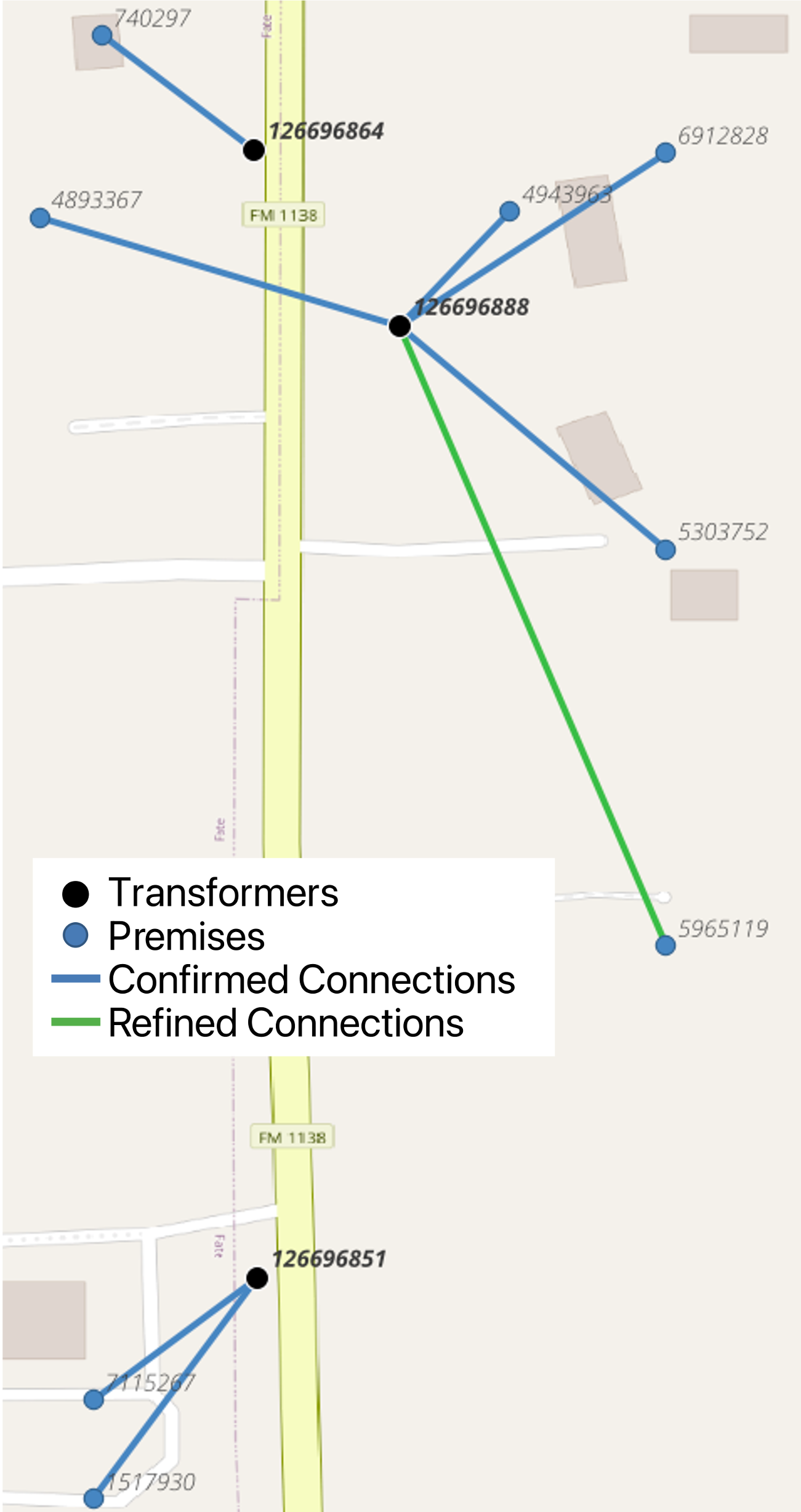}
        \caption{Refinement by our method.}
        \label{fig:case2_ours}
    \end{subfigure}
    \caption{Bad topology and refined topology in a region.}
    \label{fig:QGIS_of_Case_b}
    \vspace{-5mm}
\end{figure}

To illustrate the importance of our framework with multi-source data, we compare it to classic methods that typically utilize voltage and power measurements \cite{weng2016distributed2}. 
Purely correlation–based topology identification assumes that premises served by the \emph{same} transformer will show the strongest pair-wise voltage correlations, whereas those served by \emph{different} transformers will not. In densely built urban networks this assumption fails because adjacent transformers share extremely short secondary runs, identical conductor characteristics, and common-mode loading, which in turn produce nearly identical voltage trajectories. Figure~\ref{fig:Correlation test case} illustrates this limitation for transformers \textit{377108415} and \textit{75893988}. The recorded voltage magnitude measurements of $4$ premises, connected to the two transformers, share high similarity. The computed mutual correlations exceed $0.95$ for all pairs. Hence, voltage-based analysis can collapse the two service areas into one, mapping all premises onto a single transformer. We ran a second test to show that the same problem happens elsewhere. This test used two neighboring transformers: 126696888 and 126696851, shown in Figure \ref{fig:correlationheatmapallresult}. The right‑hand heat‑map shows all premises on 126696851. Their voltage magnitude data has the correlations to be around $0.65$. Each of those premises, however, is much more correlated with the group on 126696888. Thus, using voltage similarity may connect all premises from transformer 126696851 to transformer 126696888. The network then collapses into one big service area. Figure \ref{fig:case2_fail} shows the faulty result, compared to the original connections in the database as shown in Figure \ref{fig:case2_oncor}.

In contrast, Figure \ref{fig:case2_ours} illustrates the result of our method that employs both geographical and electrical data. Such data fusion enables the model to split the whole graph into different regions, followed by the distributed voltage-based analysis. This creates much better and robust results compared to methods that utilize all voltage data as a whole. Specifically, as described in Section \ref{shapeaware}, we compute the geographical distance in Equation \eqref{eqn:geo_dis} and the distance ratio. For transformer 126696864, we compute the distance ratio of premise 5965119 to be 5.03, which exceeds the threshold 3. Hence, the premise 5965119 is identified to be an outlier. We utilize the voltage data of the outlier and premises of the local region to do clustering, as illustrated in Section \ref{sec:recon}. Finally, the outlier premise 5965119 is correctly connected to the transformer 126696888, as shown in the green line in Figure \ref{fig:case2_ours}.

\subsection{Confidence Level Evaluations to Promote Human Trusts}


\begin{figure}[H]
\centering
\includegraphics[width=\columnwidth]{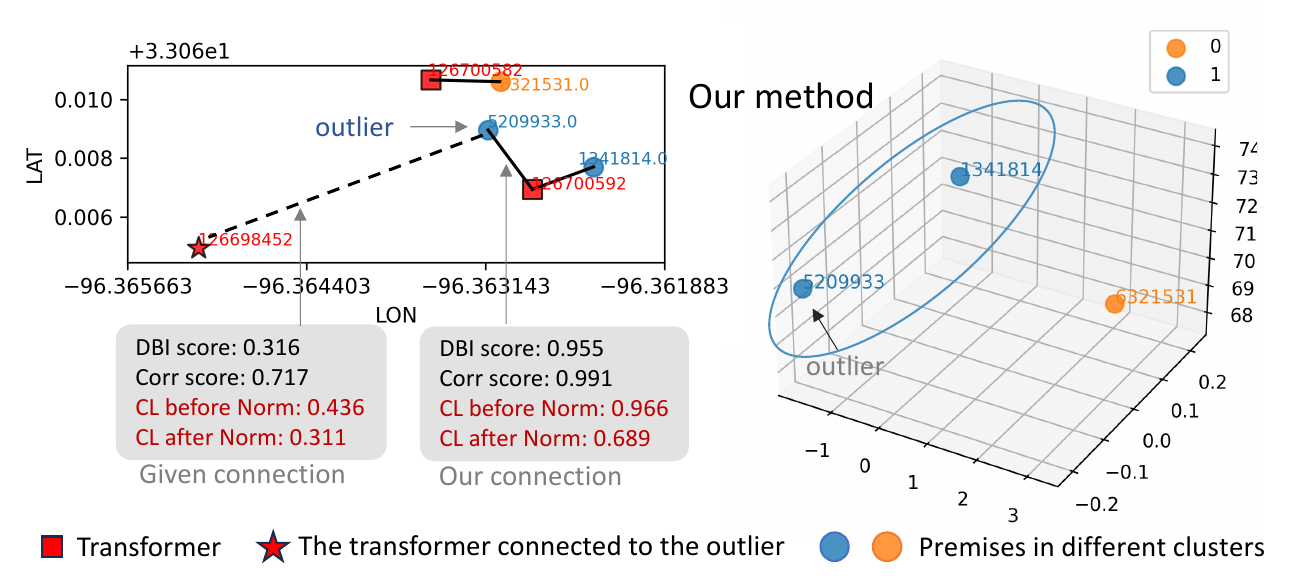}
\centering
 \caption{The confidence level designing. 
}
\label{fig:confidence_show}
\end{figure}

In this subsection, we demonstrate how to compute the confidence level to make the result trustworthy. To align the computation with human's understanding, we adopt a falsification-driven confidence metric that quantifies how much worse every \emph{alternative} connection would be. Structural separation is assessed with the Davies–Bouldin Index (DBI), whose gain relative to the best “falsified’’ alternative is converted into the sigmoid-normalised score. The process is carefully stated in Section \ref{confidencelevel}, yielding a scalar in~\([0,1]\) that depicts the algorithm's confidence about this connection. 

Figure~\ref{fig:confidence_show} contrasts the utility’s original meter–transformer mapping (“Given connection”, left) with our reassignment produced by the falsification-based confidence metric (“Our connection”, centre and right).
In the 2-D geographic inset, red squares mark transformer poles, coloured circles denote premises, and the blue star highlights the transformer currently energising the outlier. The dashed arrow shows the large spatial gap that motivated reevaluation, while the solid black arrow indicates our proposed, shorter link. Grey call-outs report the two ingredients of the confidence score: Davies–Bouldin (DBI) and correlation. Moving the outlier from its original transformer increases the DBI from 0.316 to 0.955 and the correlation from 0.717 to 0.991, resulting in a substantial rise in the overall confidence level—from 0.436 to 0.966 before normalization, and from 0.311 to 0.689 after normalization. The right-hand 3-D embedding (first two PCA components plus mean daily load) visualizes cluster geometry after reassignment: premises now lie on a coherent ellipsoid (blue envelope). The figure illustrates how the hybrid metric rewards assignments that simultaneously tighten cluster compactness and strengthen temporal similarity, yielding a higher, more trustworthy CL for operational triage.

\begin{figure*}[t!]
    \centering
    \begin{subfigure}[b]{0.48\textwidth}
        \centering
        \includegraphics[width=\textwidth]{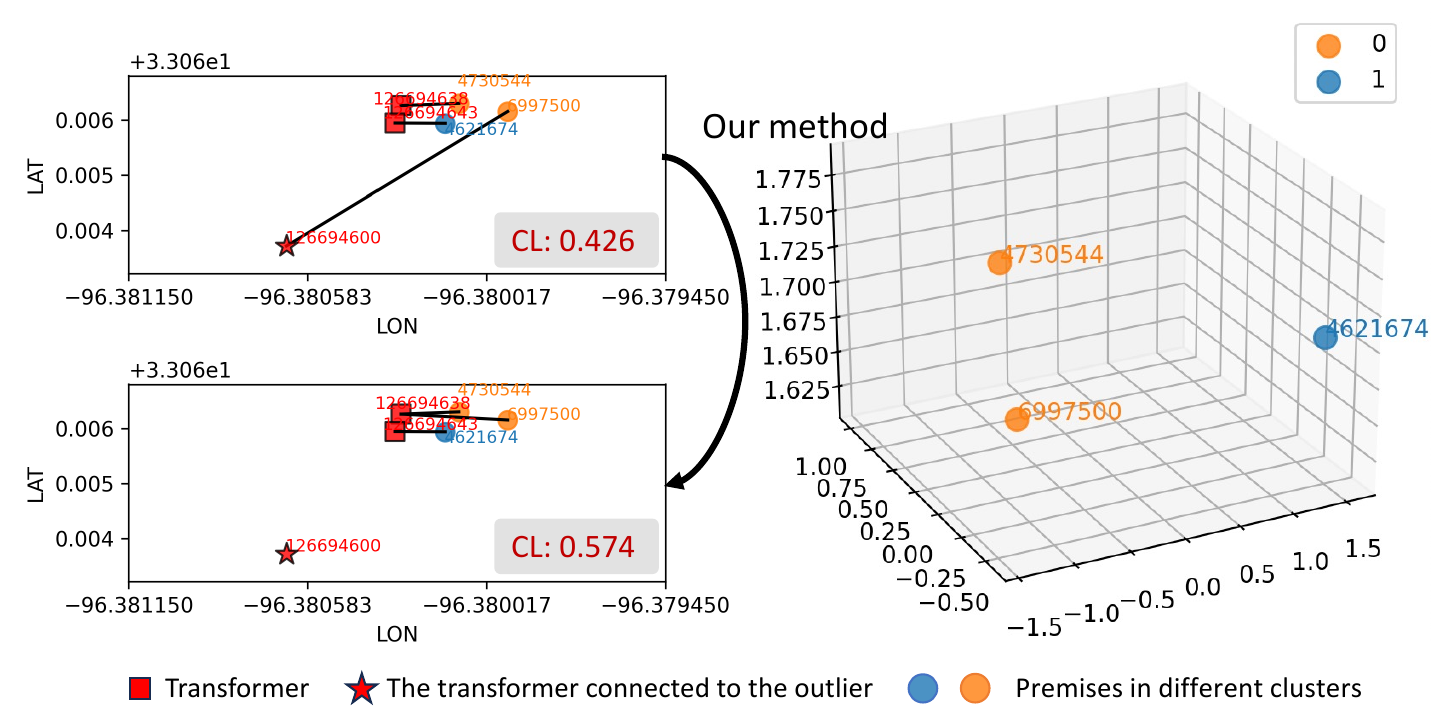}
        \caption{Results of case 1}
        \label{fig:subfig1}
    \end{subfigure}
    \hfill
    \begin{subfigure}[b]{0.48\textwidth}
        \centering
        \includegraphics[width=\textwidth]{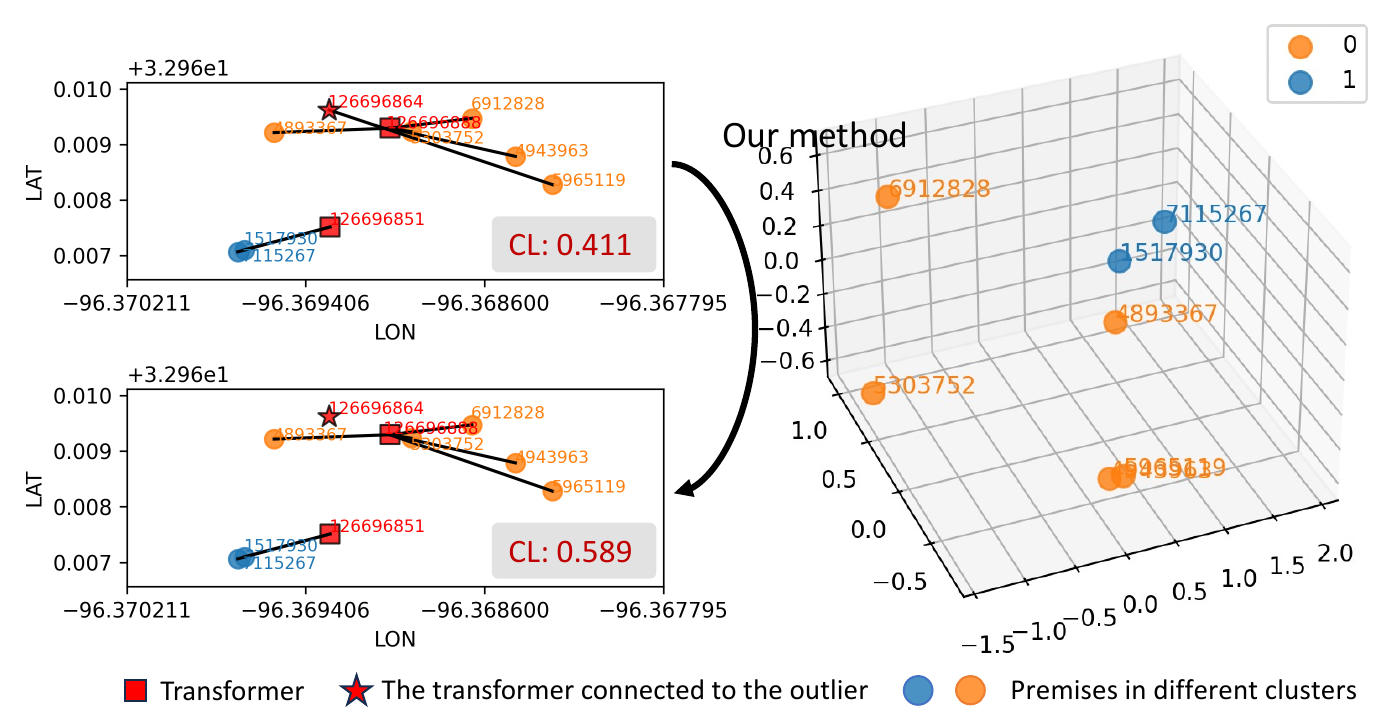}
        \caption{Results of case 2}
        \label{fig:subfig2}
    \end{subfigure}
    \\[1em]
    \begin{subfigure}[b]{0.48\textwidth}
        \centering
        \includegraphics[width=\textwidth]{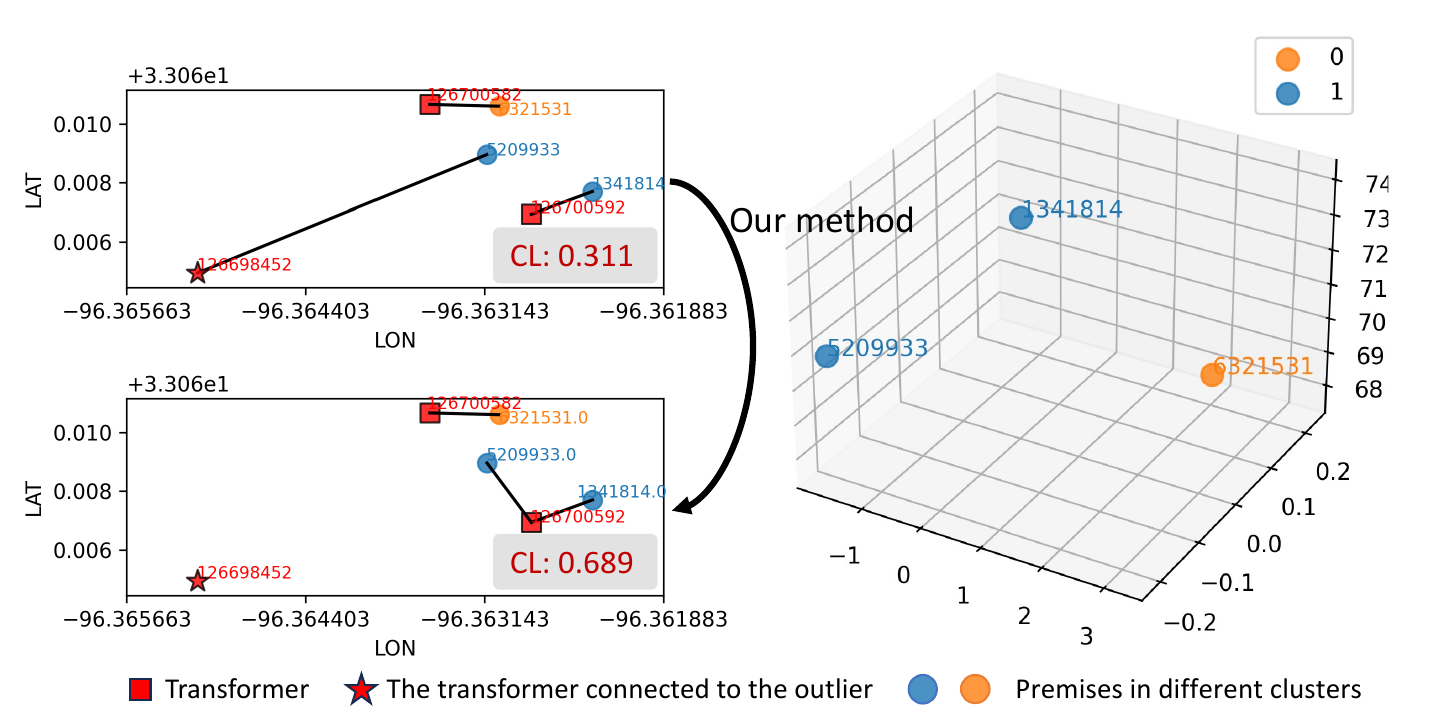}
        \caption{Results of case 3}
        \label{fig:subfig3}
    \end{subfigure}
    \hfill
    \begin{subfigure}[b]{0.48\textwidth}
        \centering
        \includegraphics[width=\textwidth]{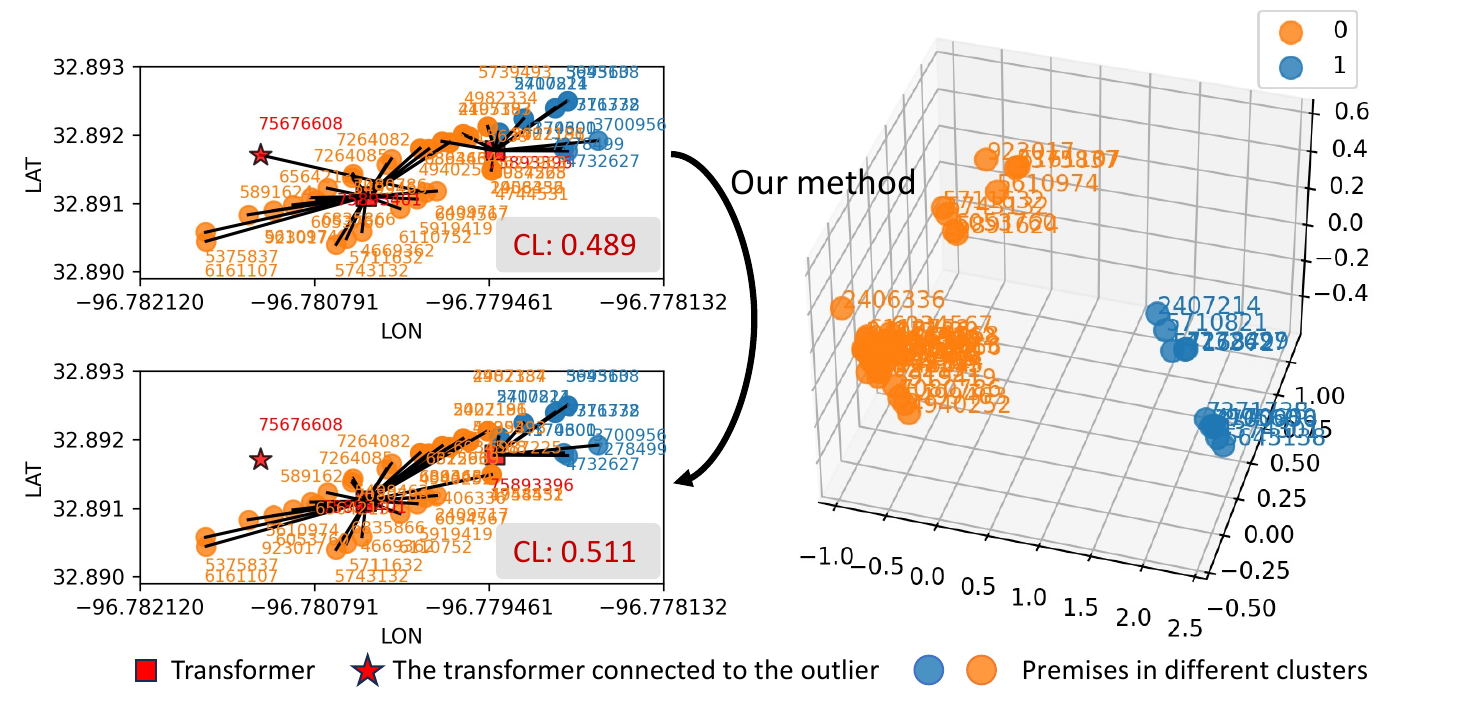}
        \caption{Results of case 4}
        \label{fig:subfig4}
    \end{subfigure}

    \caption{Four test cases with confidence level.}
    \label{fig:test case}
\end{figure*}

We validate the proposed confidence metric on four representative feeders, summarised in Figure \ref{fig:test case}. In every sub-figure, the \emph{upper inset} depicts the utility’s original meter–transformer link, whereas the \emph{lower inset} shows the reassignment recommended by our method; the grey boxes report the resulting confidence levels (CL). Across Cases 1–4 the CL rises from 0.426 → 0.574, 0.411 → 0.589, 0.311 → 0.689, and 0.489 → 0.511, respectively, demonstrating consistent improvement even when the absolute gain is modest (Case 4). These results confirm that the falsification-driven hybrid score reliably identifies and corrects low-confidence links over diverse geographic and load‐curve conditions. Figure ~\ref{fig:subfig2} shows that the greedy correlation correction achieves a confidence of only~0.411, signalling that many falsified assignments appear equally plausible. After applying our anomaly-aware geo-electrical fusion, the score rises to~0.589, and the 3-D latent embedding on the right of that figure reveals two well-separated clusters—blue versus orange—each correctly connected to its own transformer. The restored topology keeps both transformers within their thermal limits and mirrors the feeder’s physical reality.

\vspace{-3mm}

\begin{figure}[t!]
    \centering
    \includegraphics[width=0.75\columnwidth]{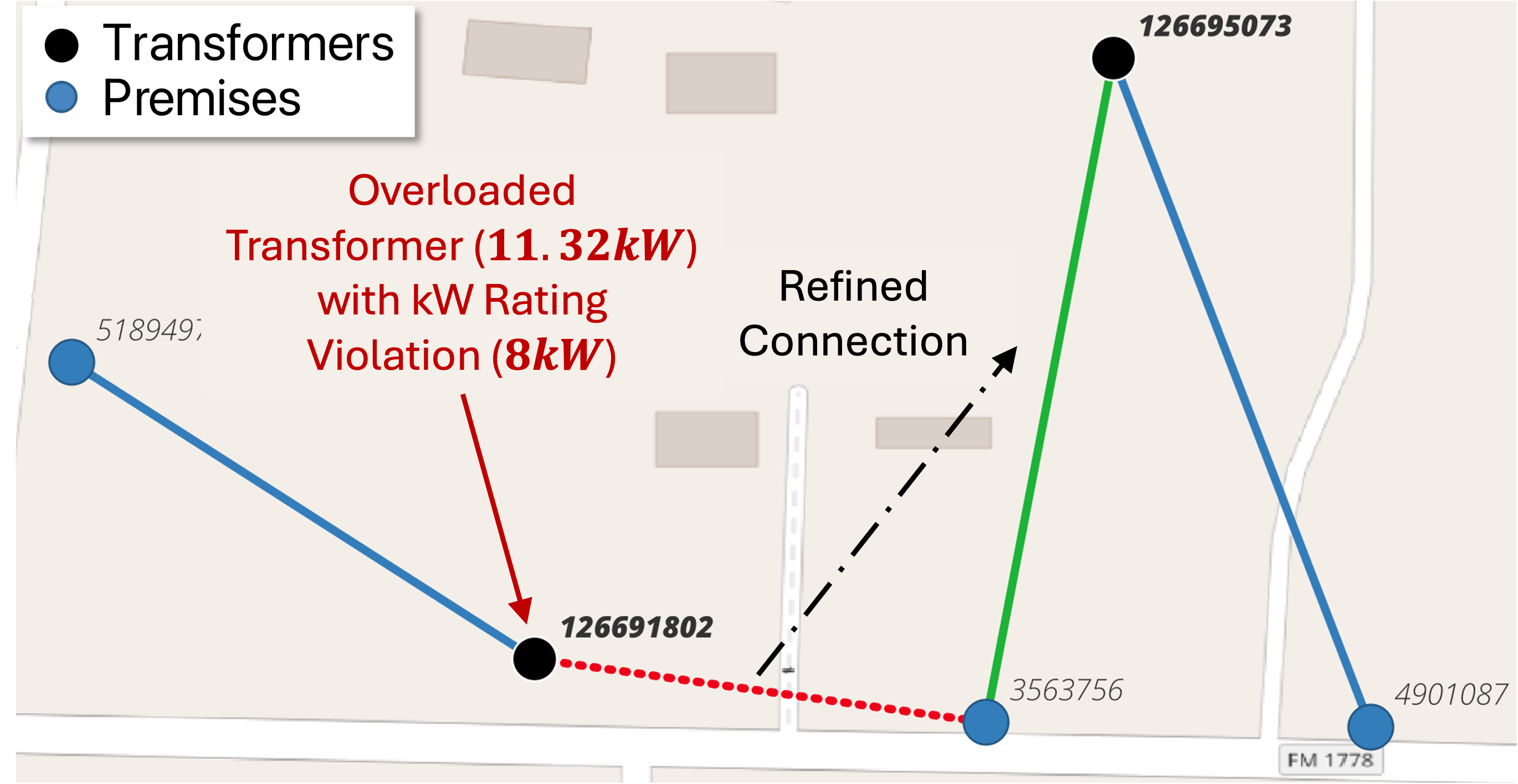}
    \caption{Refined connection by the physical constraint on one overloaded transformer (11.32 kW) with kW rating violation (8 kW).}
    \label{fig:overload_refinement}
\end{figure}

\begin{table}[h!]
    \centering
    \captionsetup{font=small}
    \caption{Example of Transformer Capacity Violations}
    \label{tab:capacity_violation}
    \footnotesize 
    \renewcommand{\arraystretch}{1.0}
    \setlength{\tabcolsep}{2.5pt} 
    \begin{tabular}{
        >{\centering\arraybackslash}p{0.17\columnwidth}|
        >{\centering\arraybackslash}p{0.2\columnwidth}|
        >{\centering\arraybackslash}p{0.18\columnwidth}|
        >{\centering\arraybackslash}p{0.18\columnwidth}|
        >{\centering\arraybackslash}p{0.18\columnwidth}
    }
        \toprule
        \textbf{ID} & \textbf{Rating (kVA)} & \textbf{Peak (kW)} & \textbf{Limit (kW)} & \textbf{Status} \\
        \midrule
        176650500  & 10  & 16.6  & 8   & Violation \\
        126691802  & 10  & 11.3  & 8   & Violation \\
        126697400  & 10  & 11.3  & 8   & Violation \\
        126695041  & 10  & 9.4   & 8   & Violation \\
        126694970  & 10  & 8.8   & 8   & Violation \\
        126700683  & 15  & 13.0  & 12  & Violation \\
        126026376  & 10  & 9.4   & 8   & Violation \\
        126026718  & 15  & 12.8  & 12  & Violation \\
        \bottomrule
    \end{tabular}
\end{table}

\subsection{Physical Constraint Validations Yield Feasible Results }



To ensure feasible topology identification, physical constraints such as transformer capacity limits are enforced. Table~\ref{tab:capacity_violation} shows an example where the total peak kilowatt (kW) usage of premises connected to a transformer exceeds its rated capacity, indicating an infeasible connection.
Case Example:
%
Transformer 126691802: The peak real power usage is 11.32~kW, while the transformer's kVA rating is 10~kVA. Assuming a power factor (pf) of 0.8, the allowed peak real power is $10~\text{kVA} \times 0.8 = 8~\text{kW}$. 
\begin{figure*}[h]
    \centering
    \begin{subfigure}[b]{0.45\textwidth}
       \centering
        \includegraphics[trim=0pt 5pt 0pt 5pt, clip, width=\textwidth]{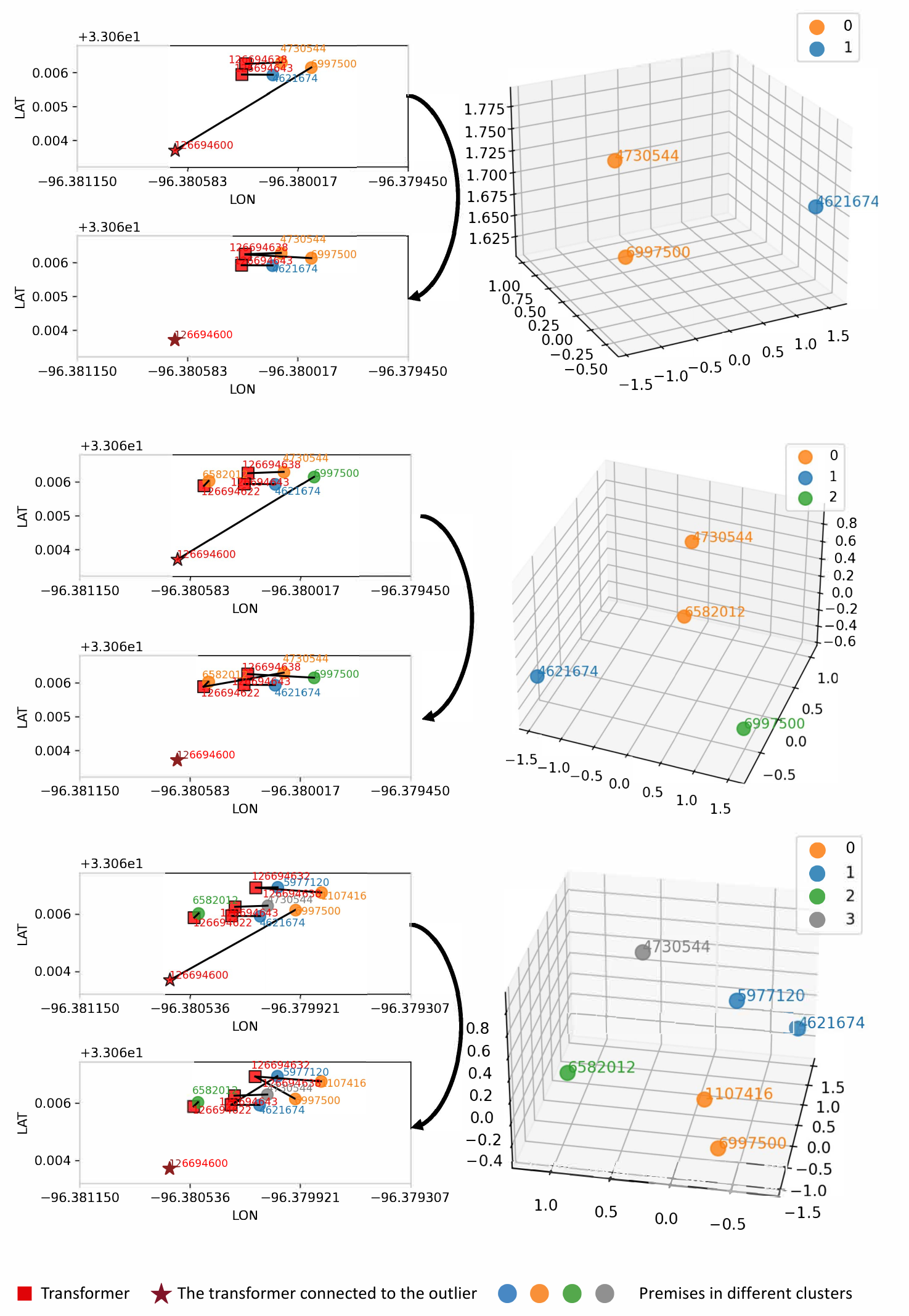}
        \caption{Sensitivity analysis on region 1.}
        \label{fig:sensitivity_case1}
    \end{subfigure}
    \hfill
    \begin{subfigure}[b]{0.45\textwidth}
        \centering
        \includegraphics[trim=0pt 5pt 0pt 5pt, clip, width=\textwidth]{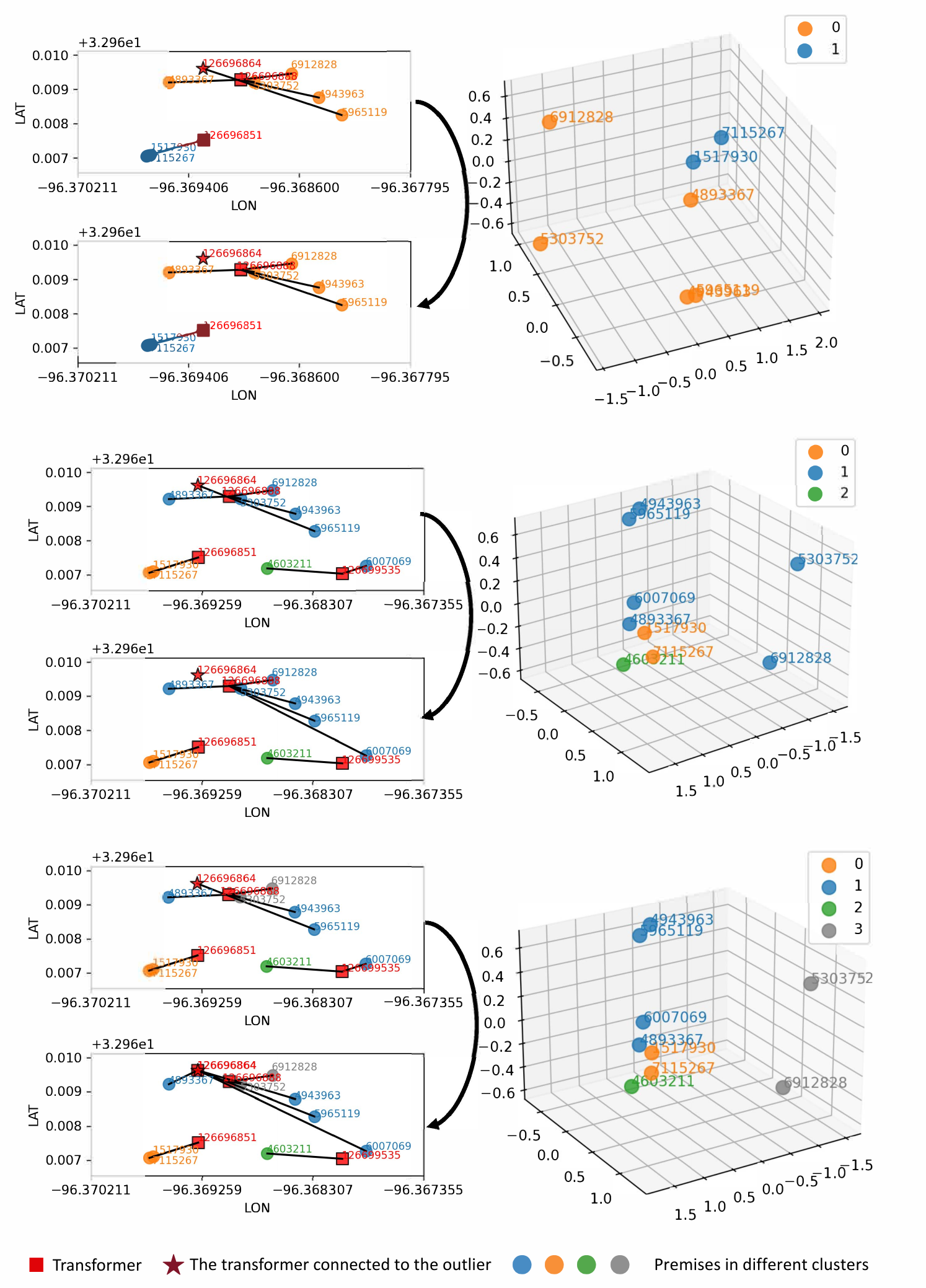}
       \caption{Sensitivity analysis on region 2.}
        \label{fig:sensitivity_case2}
    \end{subfigure}
    \\[-0em]
    \begin{subfigure}[b]{0.45\textwidth}
        \centering
        \includegraphics[trim=0pt 5pt 0pt 5pt, clip, width=\textwidth]{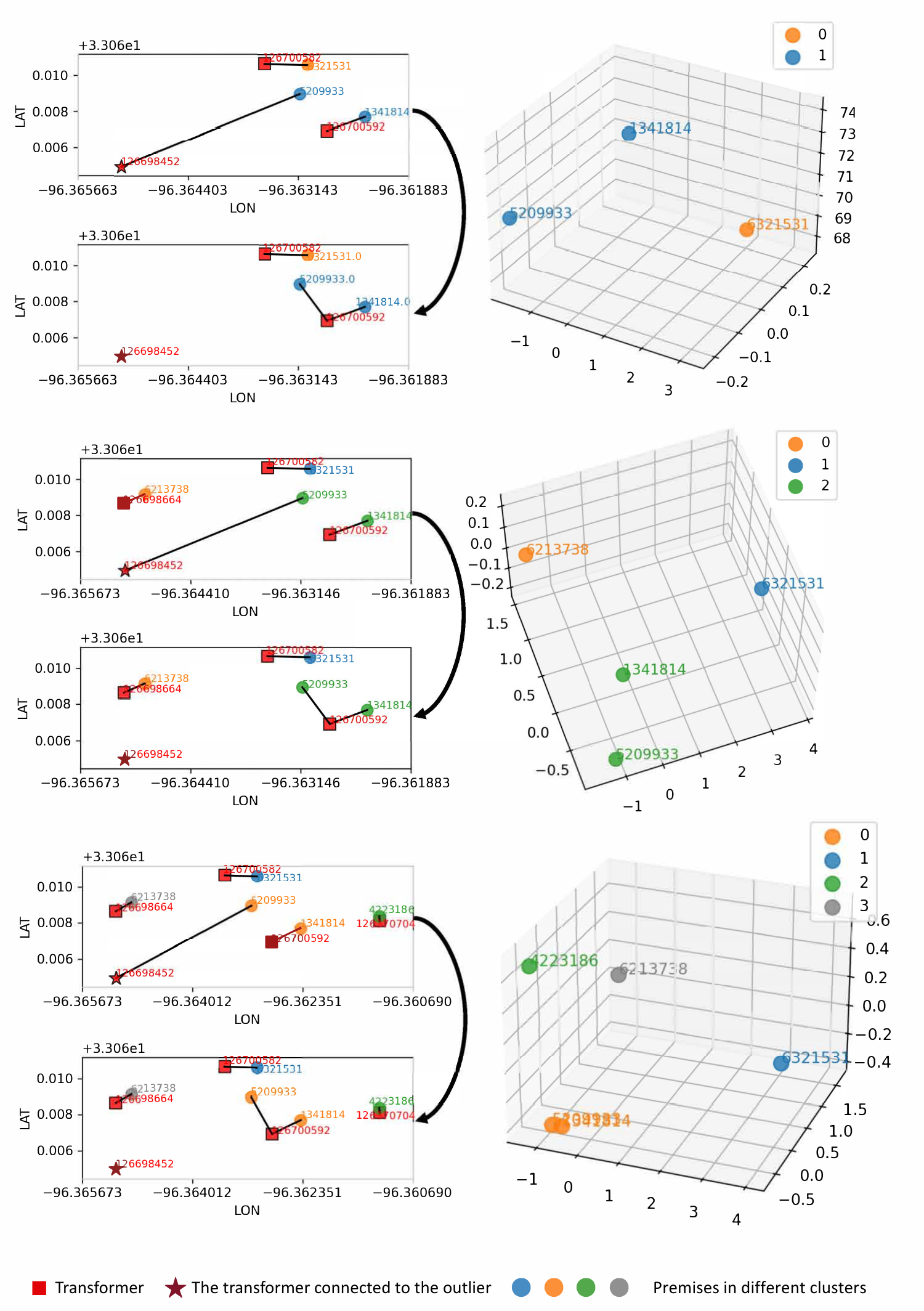}
       \caption{Sensitivity analysis on region 3.}
        \label{fig:sensitivity_case3}
    \end{subfigure}
    \hfill
    \begin{subfigure}[b]{0.45\textwidth}
       \centering
        \includegraphics[trim=0pt 5pt 0pt 5pt, clip, width=\textwidth]{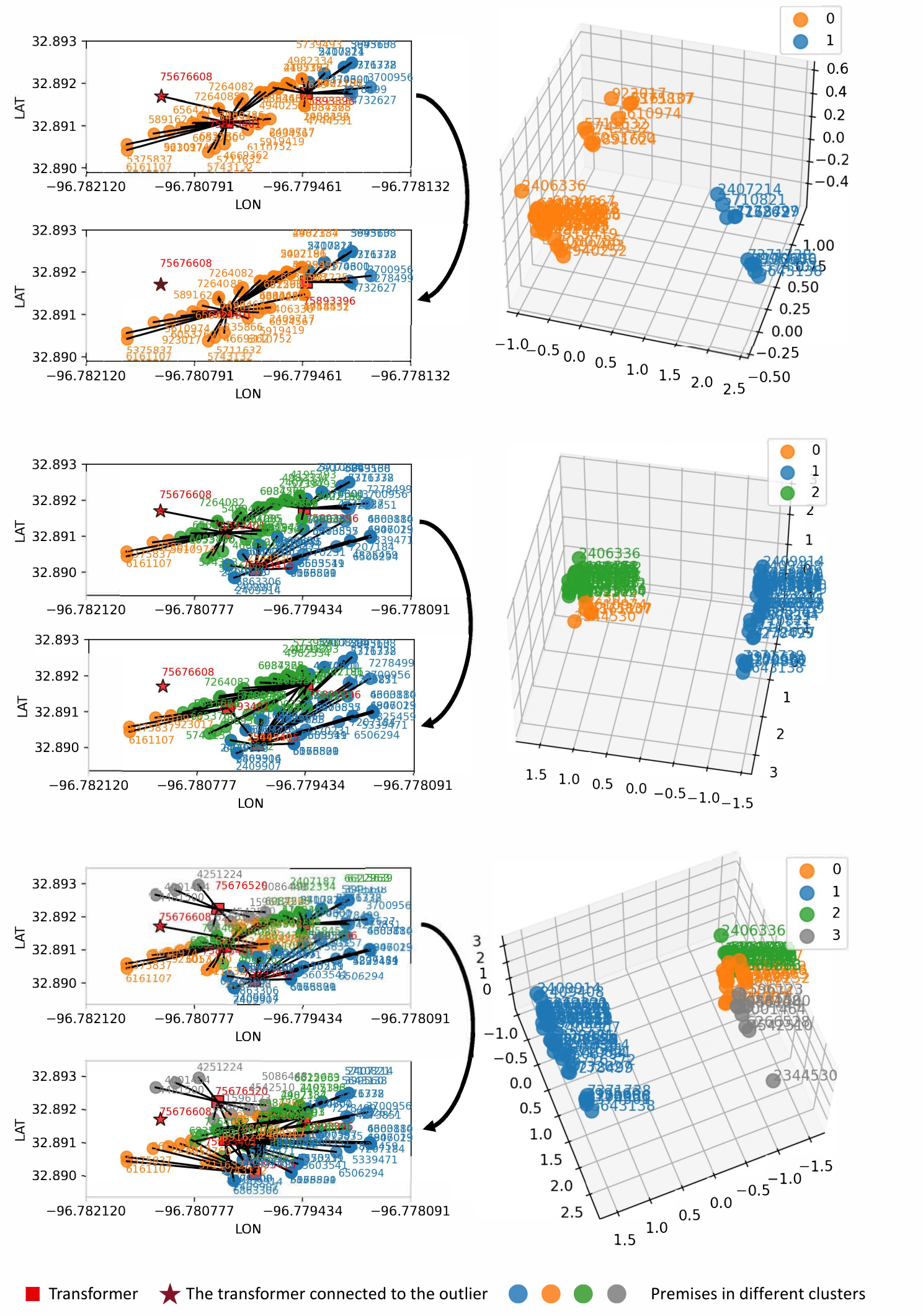}
        \caption{Sensitivity analysis on region 4.}
        \label{fig:sensitivity_case4}
    \end{subfigure}
    \caption{Sensitivity analysis on four regions.}
    \label{fig:sensitivity analysis}
\end{figure*}
\clearpage
The observed peak of 11.32~kW therefore exceeds the limit, resulting in a violation. The refined connection is shown in Figure \ref{fig:overload_refinement}.

The general condition for identifying capacity violations is:
$P_{\text{peak}} > \text{kVA}_{\text{rated}} \times \text{pf}$.
To solve the overloading issue, we have a validation step. Figure \ref{fig:overload_refinement} shows before and after the refinement performed on transformer 126691802. After the refinement, there is no overloading on the transformer anymore, and the wrong connection is corrected as well.

\vspace{-3mm}

\subsection{Sensitivity Analysis}

In this subsection, we report the sensitivity analysis with respect to the number of assigned transformers for an outlier, which is used in our clustering algorithm to reconnect the outlier to one of the transformers. Basically, it's the K value in K-means clustering in Section \ref{sec:recon}. Table \ref{tab:outlier_assignments} shows the transformer assignment results across four regions. A sensitivity analysis reveals the robustness of our model under different configurations: The model maintains the same results when the number of transformer candidates is 2 or 3 in all four regions. Extensive figures and restuls are shown in Fig. \ref{fig:sensitivity analysis}. However, when the number of candidates increases to 4, significant changes in the assignment results are observed, especially in Regions 1 and 4. These findings suggest that the model's performance and consistency are influenced by the size of the region and the number of candidate transformers. We recommend making $K=2$ and running the model in relatively small regions to minimize variations. 

\begin{table}[ht]
\centering
\caption{The transformer assignment (TA) results across four regions.}
\begin{tabular}{c|c|c|c|c}
\toprule

 & \textbf{Region 1} & \textbf{Region 2} & \textbf{Region 3} & \textbf{Region 4} \\
\cline{1-5}
Outlier & 6997500 & 5965119 & 5209933 & 6034567 \\
\hline
TA (\textbf{K=2}) 
 & 126694638 & 126696888 & 126700592 & 75893401 \\
\hline
TA (\textbf{K=3}) 
 & 126694638 & 126696888 & 126700592 & 75893401 \\
\hline
TA (\textbf{K=4}) 
 & 126694632 & 126696864 & 126700592 & 75676520 \\
\bottomrule
\end{tabular}
\label{tab:outlier_assignments}
\end{table}

\vspace{-5mm}

\section{Conclusion and Future Work}
\label{sec:conclu}

%
This paper presented a scalable framework for topology identification in distribution grids, developed in collaboration with Oncor. By combining constraint-guided learning and multi-source data fusion, the approach addresses real-world utility data challenges such as noise, missing labels, and inconsistent metadata. The framework introduces confidence-aware inference that quantifies reliability and enables selective validation, enhancing trust and efficiency in field operations.
Validated on 3 feeders of AMI meter data, the method achieved over 95\% accuracy in topology reconstruction, while reducing computation time and improving phase assignment robustness. Future work will focus on applying and testing the designed workflow on Oncor's grid-wise connectivity model. Additional goals include improving performance in sparsely labeled regions and extending the framework to support real-time updates and long-term adaptability in dynamic grid environments.

\vspace{-3mm}
\section{Acknowledgments}
The authors would like to thank Kai Asberry (Oncor) for assistance with data preparation, Jeremy Williams (Oncor) for insightful feedback throughout the project, and Xiyun Wang (Oncor) for help with code revisions.

\label{sec:conclu}
\bibliographystyle{IEEEtran}
\bibliography{IEEEabrv,reference}

\end{document}